\pdfoutput=1

\documentclass[11pt]{article}

\usepackage[final]{acl}

\usepackage{times}
\usepackage{latexsym}
\usepackage{float}
\usepackage{subcaption}
\usepackage{listings}
\usepackage{amsmath}
\usepackage[T1]{fontenc}

\usepackage[utf8]{inputenc}

\usepackage{microtype}

\usepackage{inconsolata}

\usepackage{graphicx}
\title{Navigate Complex Physical Worlds via Geometrically Constrained LLM}

\author{
    \textbf{Yongqiang Huang\textsuperscript{1}}, 
    \textbf{Wentao Ye\textsuperscript{2}}, 
    \textbf{Liyao Li\textsuperscript{2}}, 
    \textbf{Junbo Zhao\textsuperscript{2,*}} \\[5pt]
    \textsuperscript{1}College of Energy Engineering, Zhejiang University\\ 
    \textsuperscript{2}College of Computer Science and Technology, Zhejiang University\\
    {Emails: \{hyq.cee, yewt01, liliyao, j.zhao\}@zju.edu.cn} \\[5pt]
    {\textbf{* Corresponding author:} \href{mailto:j.zhao@zju.edu.cn}{j.zhao@zju.edu.cn}}
}

\begin{document}
\maketitle

\begin{abstract}
This study investigates the potential of Large Language Models (LLMs) for reconstructing and constructing the physical world solely based on textual knowledge. It explores the impact of model performance on spatial understanding abilities. To enhance the comprehension of geometric and spatial relationships in the complex physical world, the study introduces a set of geometric conventions and develops a workflow based on multi-layer graphs and multi-agent system frameworks. It examines how LLMs achieve multi-step and multi-objective geometric inference in a spatial environment using multi-layer graphs under unified geometric conventions. Additionally, the study employs a genetic algorithm, inspired by large-scale model knowledge, to solve geometric constraint problems. In summary, this work innovatively explores the feasibility of using text-based LLMs as physical world builders and designs a workflow to enhance their capabilities.
\end{abstract}

\section{Introduction}
LLMs acquire extensive world knowledge embedded in textual data through pre-training. This raises an intriguing question: can LLMs reconstruct and simulate the physical world using this textual knowledge? The physical world, characterized by complex geometric and physical constraints, can be abstracted into fundamental geometric shapes. Utilizing a custom-designed engine, we simplify the 3D world’s geometric content into basic cube combinations. This work pioneers the exploration of text-only LLMs as potential builders of the physical world, leveraging their pre-trained knowledge to understand and generate 3D spatial representations purely from textual descriptions.

Some preliminary work on world-building has explored constructing 3D worlds at the image level. Techniques like 3D-VAE-GAN \citep{wu2016learning} and Pix2Vox \cite{xie2019pix2vox} combine Variational Autoencoders (VAEs) \citep{kingma2013auto} and Generative Adversarial Networks (GANs) \cite{goodfellow2020generative} to generate high-quality 3D models with precise shape and pose control. AtlasNet \citep{groueix2018papier} approximates 3D surfaces by learning a set of 2D textures, effectively handling irregular topologies. Despite their impressive quality, these models struggle with simulating complex physical interactions and maintaining spatial consistency due to intricate and dynamic geometric constraints\cite{li2024advances}.

Some methods rely on high-precision geometric libraries or external knowledge bases for human-level prior knowledge. For instance, \citet{sun20233d} and \citet{zhou2024scenex} use LLMs to generate 3D scene images by calling Blender APIs based on user requirements. \citet{wu2024external} proposes combining external knowledge bases to generate 3D scenes from sketches. However, these methods heavily depend on external libraries and interfaces, which lack flexibility and face challenges like resource maintenance, copyright disputes, and network security issues\cite{gao2014active}.

We explored how to leverage LLM pre-training knowledge to autonomously guide complex geometric constraints. Our evaluation compared the spatial construction and geometric relationship understanding abilities of GPT-3.5-turbo and GPT-4, revealing that GPT-4 excels in spatial construction tasks due to its superior performance. we also introduced an innovative multi-agent approach for 3D scene construction, establishing geometric conventions at three levels (center, axis, and surface) to standardize the spatial relationships of 3D objects as understood by LLMs. This multi-level graph-driven approach enhances the spatial understanding and reasoning capabilities of LLMs. The workflow ensures information consistency and uniformity, mitigating data silos and redundancy issues, while enabling LLMs to explore their ability to understand geometric relationships of physical world.
\begin{figure*}[h]
    \centering
    \includegraphics[width=1\linewidth]{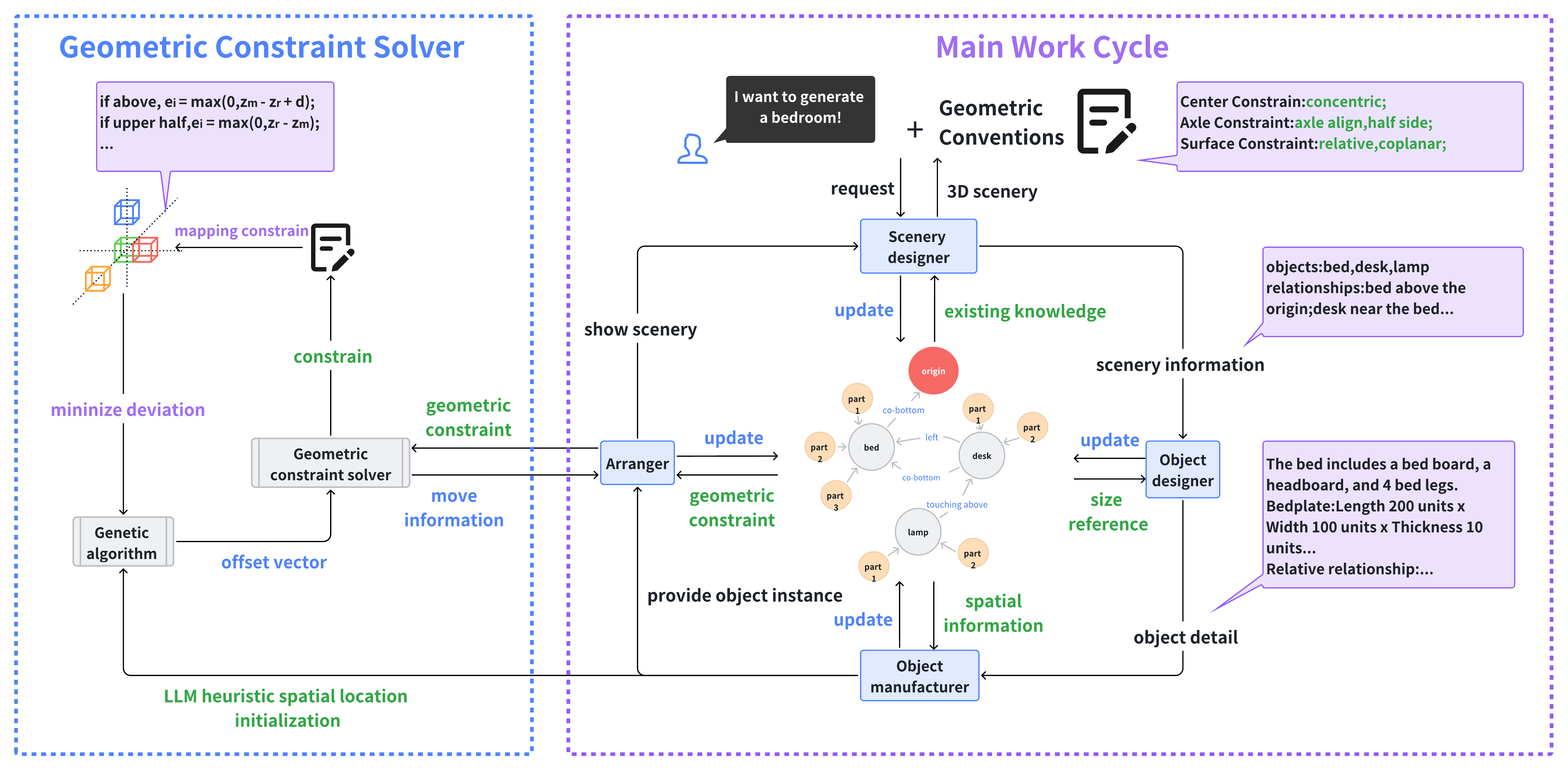}
    \caption{The entire workflow is based on geometric conventions and relies on multiple agents to carry out 3D scene construction work around the graph. The user's demand information will be refined layer by layer by designers and used to generate object instances. Finally, the arranger will use the mapping from geometric constraints to deviations and a genetic algorithm solver to determine the correct placement position of the object.}
    \label{fig:workflow}
\end{figure*}
\section{Related Work}

\subsection{Generation Based On 3D Graphics}
The application of GANs and VAEs in 3D scene generation has made notable progress in recent years. \citet{chan2022efficient} provides a method which can synthesize high-resolution, multi-view consistent images in real-time and also generate high-quality 3D geometry. \citet{xie2019pix2vox} proposes a context-aware convolutional neural network to reconstruct 3D voxel models from single and multi-view images. This method uses GANs to enhance the detail and structural accuracy of the generated 3D models. \citet{wu2016learning} combines GAN for generating and controlling 3D objects, producing high-quality 3D models with shape control. \citet{groueix2018papier} introduces a 3D surface generation method by learning a collection of 2D maps to approximate 3D surfaces, handling irregular topologies.Besides, \citet{tang2024cycle3dhighqualityconsistentimageto3d} find a method to use 2D diffusion model which can further control the generated content and inject reference-view information for unseen views.

These works typically offer high quality and realism, creativity, and diversity in generated content. However, they also face challenges such as high data dependency, complexity, and computational intensity.Moreover, such work often overlooks the complex geometric relationships between objects in the physical world.

\subsection{Generation Based On External Libraries}
The quality and availability of numerous 3D models have significantly improved. \citet{tang2024cycle3dhighqualityconsistentimageto3d} provide a large amount of 3D materials. And \citet{zhou2018open3dmodernlibrary3d} provide an open-source library that supports rapid development of software for processing 3D data.It benefits research that utilizes LLMs to invoke open-source models and achieve scene graph construction. \citet{sun20233d}, based on a multi-agent system, call the Blender interface to generate 3D scene images according to user requirements. SceneX \citep{zhou2024scenex} employs LLMs to drive procedural modeling, utilizing Blender APIs and a vast array of procedural assets. \citet{wu2024external} offer an approach that combines user sketches with external knowledge, progressively generating 3D scenes through a scene diffusion model. Their work demonstrates how these agents can leverage external tools and model libraries to automate the construction and understanding of scene graphs.

Utilizing existing model libraries offers significant advantages in terms of efficiency, scalability, and flexibility in scene generation. However, due to the heavy reliance on external libraries and external materials, the work in question exhibits inconsistent material quality, poses high maintenance complexity, demonstrates insufficient flexibility, and involves copyright challenges.

\section{Method}
\subsection{Graph Runs Through the Entire Workflow}
Multi-agent systems have demonstrated effective performance in segmenting complex problems into numerous sub-problems and resolving them \citep{grossi2023advances}, aligning with the step-by-step decomposition of three-dimensional scene concepts and the meticulous refinement of generated content at each stage in this work. And implementing information alignment between proxy groups is a huge challenge\citep{han2024llmmultiagentsystemschallenges}. Inspired by \citet{Qi_2023} and \citet{ranasinghe2024learninglocalizeobjectsimproves}, we choose graph database as the medium. In our work, we use GPT-4 \citep{openai2023gpt4} as the basis for the agent and Neo4j \citep{neo4j} database to store our graph. By employing a graph database to capture spatial information and representing shapes and their geometric relationships with nodes and edges, complex geometric relationships can be managed flexibly. The graph database records scene information, providing a comprehensive overview of user objectives and scene graphs throughout the workflow. This ensures that generated scenes align with predefined spatial constraints and design specifications by integrating relational processing with large model generation capabilities, offering a flexible and efficient solution for managing complex spatial data and scene generation.

\subsubsection{Scenery Designer}
Graph databases can stably and comprehensively record object information in existing scenes, thereby reducing scene graph generation errors caused by illusions or memory problems in LLMs, such as reconstructing existing objects or using non-existent objects as reference points. By providing detailed scene information to LLMs, the graphics database helps to develop plans that are consistent with the given semantics and do not conflict with the current scene graph. Based on this, the scene designer will mobilize their internal world knowledge to design a scene that is semantically consistent with the input, including the main objects in the scene and the spatial geometric relationships between objects.

\subsubsection{Object Designer}
After the scene planning is completed, the object designer needs to design objects with appropriate structure and size based on the existing reference objects in the scene. On the one hand, image databases are needed to provide background information, and on the other hand, LLMs themselves require a certain level of common sense knowledge and reasoning ability to lay a more detailed foundation for the next step of object creation.

\subsubsection{Object Manufacturer}
After completing the object design phase, we proceed to the construction phase. At this stage, LLMs require a thorough understanding of the descriptive statements used by object designers, particularly those describing the interrelationships between internal modules of the object. This ensures alignment between the generated objects and their descriptive statements. We have observed that models with weaker performance, such as the GPT-3.5 turbo\cite{openai2023gpt35turbo}, often have poor performance in this step, regardless of the level of detail provided by the designer.
Additionally, to minimize the risk of spatial divergence when using genetic algorithms in later permutation calculations, the initial position of the object should be proximate to its main reference object, typically adhering to their relative spatial relationships. Here, a graphical database becomes crucial, as it offers detailed information about the size and position of reference objects, as well as their approximate relative relationships. This information is essential to guide LLMs in utilizing their internal knowledge effectively.

\subsubsection{Arranger}
Following the construction of the object, further optimization of its spatial position is required to meet specific spatial requirements, such as those related to smaller particle sizes. Initially, the relationship information between the newly constructed object and the reference object must be extracted from the graph database. This information is then used to perform further inference and to supplement any missing spatial constraints. Based on these completed spatial constraints, the appropriate constraint equations can be selected for positional optimization.

The graph database provides a comprehensive understanding of global scene information at each layer of the workflow and provides necessary information for each layer to complete tasks. It can efficiently manage complex relationships and dependencies, enabling each level to accurately locate and process relevant information in complex scenarios. 

\subsection{Geometric Conventions}
Inspired by the work of \citet{hedau20113d} and \citet{klein1998role}, we recognize that clearly and systematically representing the relative positions of objects in space is beneficial for enhancing the spatial reasoning capabilities of LLMs. Consequently, we have devised a spatial convention that encompasses three levels of constraint relationships: geometric center, axis, and surface, with varying degrees of constraint strength.By integrating different spatial conventions, we can flexibly and accurately determine the positions of objects within a reasonable range. This set of spatial conventions is integral to our entire workflow. Through the implementation of a unified spatial convention system, we ensure consistency and standardization throughout the workflow.

An example of the spatial convention we designed is as follows:
\subsubsection{Geometric Center Relationship Constrain}
\begin{itemize}
    \item Concentric relationship:
       \begin{equation}
            x^{c}_{m} = x^{c}_{r}, \quad y^{c}_{m} = y^{c}_{r} \quad \text{and} \quad z^{c}_{m} = z^{c}_{r}
        \end{equation}
\end{itemize}

\subsubsection{Axle Relationship Constraint}
\begin{itemize}
    \item x align:
        \begin{equation}
            x^{c}_{m} = x^{c}_{r}
        \end{equation}
    \item front half:
        \begin{equation}
            x^{c}_{\text{r}} > x^{c}_{\text{m}}
        \end{equation}
\end{itemize}
\subsubsection{Surface Relationship Constraint}
\begin{itemize}
    \item front:
        \begin{equation}
            x^{b}_{r}- x^{f}_{m} = d
        \end{equation}

    \item coplanar front:
        \begin{equation}
            x^{t}_{r} = x^{t}_{m}
        \end{equation}

\end{itemize}
To avoid misunderstandings, we briefly declare the following symbols:
\begin{itemize}
     \item x, y and z represent the projections of the corresponding parts of the object on that axis
     
    \item In superscripts, f, b and t, etc. respectively represent the corresponding surfaces of the object, such as the front, back/bottom, and top surfaces. And c represents the geometric center.
    
    \item In the subscript, r and m represent the reference object and the object to be moved, respectively. And d stands for distance.
\end{itemize}

\subsection{Graph Driven LLM Spatial Inference}
The final layer of the workflow is called the arranger, responsible for the spatial arrangement of generated objects in the scene. \citet{wei2024construct}discussed Detailed introduction on how to construct a knowledge graph of geographic spatial data, as well as how to express and infer spatial relationships. Inspired by this, this work maps the relative positional relationships of objects to a graphics database. By setting strong and weak reference objects, we provide different levels of constraints for the object to be moved. With the continuous enrichment of graphic information, our framework will provide increasingly accurate spatial constraints. After determining the spatial constraints, the LLM inspired genetic algorithm is used to solve the spatial constraints, which is used to update the spatial position of the object to be moved and dynamically update the graphic data. This layer utilizes a graphical database to store entities and their spatial relationships, establishing and updating spatial constraints at the granularity of objects. The process specifically includes several steps:

\subsubsection{Graph Database Interaction}
Arranger interacts with graphical databases to generate more detailed relationship information and select the correct constraint equation according to it.  Based on the provided rough relationship pairs, the arranger select the strong reference object which will provides 1 to 3 constrains from the graph database and return the weak reference objects which provides 0 to 2 constrains and be associated with the strong reference object. In this way, the computational complexity of constraints can be reduced. The LLM agent will obtain various types of information about the reference object, including its dimensions and spatial positions. It will then infer and add new spatial constraints within the basic spatial constraint framework and select the correct constraint equation for genetic algorithm calculation of accurate spatial positioning.

\subsubsection{Genetic Algorithm for Solving Geometric Relationships}
Given the global optimization capabilities of the genetic algorithm and its effective use with heuristic initialization, we ultimately opted for the genetic algorithm to address the spatial constraints.
When LLM completes spatial constraints and selects the correct geometric equation, the permutator pass the parameters to the genetic algorithm\citep{shapiro1999genetic} solver to optimize the geometric relationships and  further adjust and update the spatial position of the objects initialized by LLM.

Each object is composed of multiple blocks, with each block represented by its centroid coordinates and three-dimensional dimensions. The specific representation is as follows:

Single block representation:

\[
b_i = \{c_i, d_{i1}, d_{i2}, d_{i3}\}
\]

where \(c_i = (x_i, y_i, z_i)\) is the centroid coordinates, and \(d_{i1}, d_{i2}, d_{i3}\) represent the length, width, and height, respectively.

Object representation:

\[
O_i = \{b_{i1}, b_{i2}, \ldots, b_{in}\}
\]

where \(O_i\) represents an object composed of multiple blocks \(b_{ij}\).
In addition, the spatial information of objects can also be represented as follows:

\[
O_i = \{C_i,D_{i1}, D_{i2}, D_{i3}\}
\]

where \(C_i\) is the centroid coordinates, and \(D_{i1}, D_{i2}, D_{i3}\) represent the length, width, and height of \(O_i\) respectively.\\

We define various types of spatial constraints to describe the relative spatial relationships between objects. Below are examples of above, and upper half:
\[
\text{above} : z^{b}_{m} \geq z^{t}_{r} + d
\]
\[
\text{upper half} : z^{c}_{m}\geq z^{c}_{r}
\]
To generate appropriate constraint equations, we abstract the reference object as a block and generate movable object pairs with reference part relationships for each object. Then, based on the generated relationship pairs, we generate appropriate constraint equations and pass them to the genetic algorithm for solution.

Assume we have multiple reference objects \(R_k\) and a movable object \(M\), each pair \((R_k, \text{relation}, M)\) can be represented as a set of constraint formations:
\begin{equation}
\begin{split}
e_i = 
\begin{cases}
\max \left(0, z^{c}_{m} - z^{c}_{r} + d\right),\text{if above} \\
\max \left(0, z^{c}_{r} - z^{c}_{m}\right), \text{if upper half}
\end{cases}
\end{split}
\end{equation}
The optimization goal is to minimize the total error:
\[
\min{E} = \min{\sum_{i=1}^{N} e_i^2}
\]
To determine effective motion vectors, we employed a genetic algorithm inspired by LLM initialization. Objects are generated at specific positions based on global and reference content, partially fulfilling constraint requirements. The algorithm's initialization is then refined based on the size of both the reference object and the object to be moved, enhancing the optimization process. Each genome consists of three XYZ coordinates representing motion vectors. The total error \(E\) of each individual is calculated to assess fitness, with top-performing individuals selected for crossover and mutation. During crossover, parent DNA combines to produce new offspring, and mutations make fine adjustments to coordinates. This process iterates until a set number of generations or error convergence is achieved, gradually approaching the optimal solution.

\section{Experiment}
In this section, we will discuss the factors affecting the quality of the 3D scene graph generated by the LLM from two aspects. The first influencing factor is the model's ability. We test the generation performance of the base models GPT-3.5-Turbo and GPT-4 without using the framework. The second influencing factor is the degree of integration with the work framework. We set up three sets of experiments to explore the complete use of the work framework, including ablation experiments to analyze the impact of removing certain components.
\subsection{Model Performance Impact}
Our experiment found a strong correlation between LLM performance and spatial understanding. Evaluating GPT-3.5-Turbo and GPT-4-0125 on object and scene generation tasks, we observed that GPT-3.5 had poor spatial comprehension and simplistic outputs. In contrast, GPT-4 showed improved spatial concepts and multi-object scene generation but still used simple blocks with limited detail.
\subsubsection{Object Generation}
\begin{figure}[H]
    \centering
    \begin{subfigure}[b]{0.18\textwidth}
        \centering
        \includegraphics[width=\textwidth]{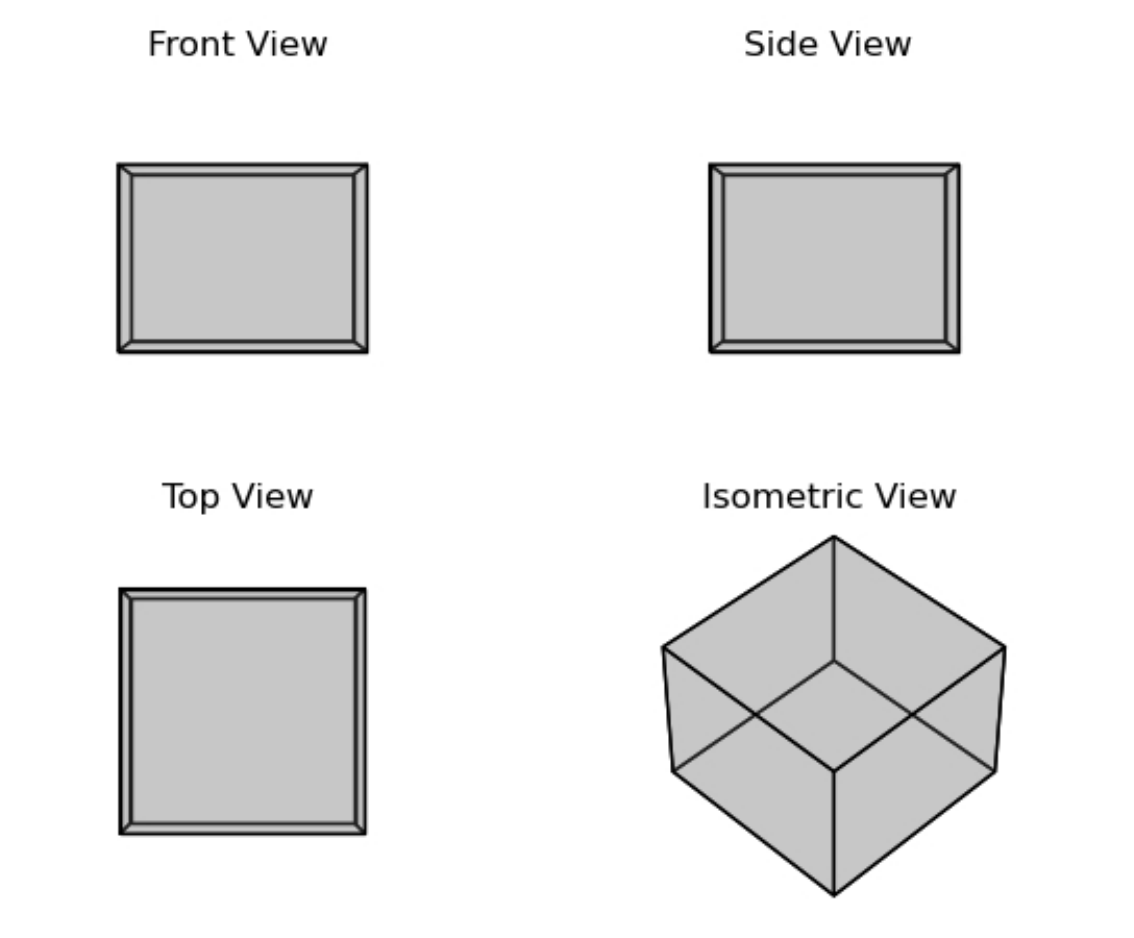}
        \caption{Chair(3.5)}
        \label{fig:chair3}
    \end{subfigure}
    \hfill
    \begin{subfigure}[b]{0.18\textwidth}
        \centering
        \includegraphics[width=\textwidth]{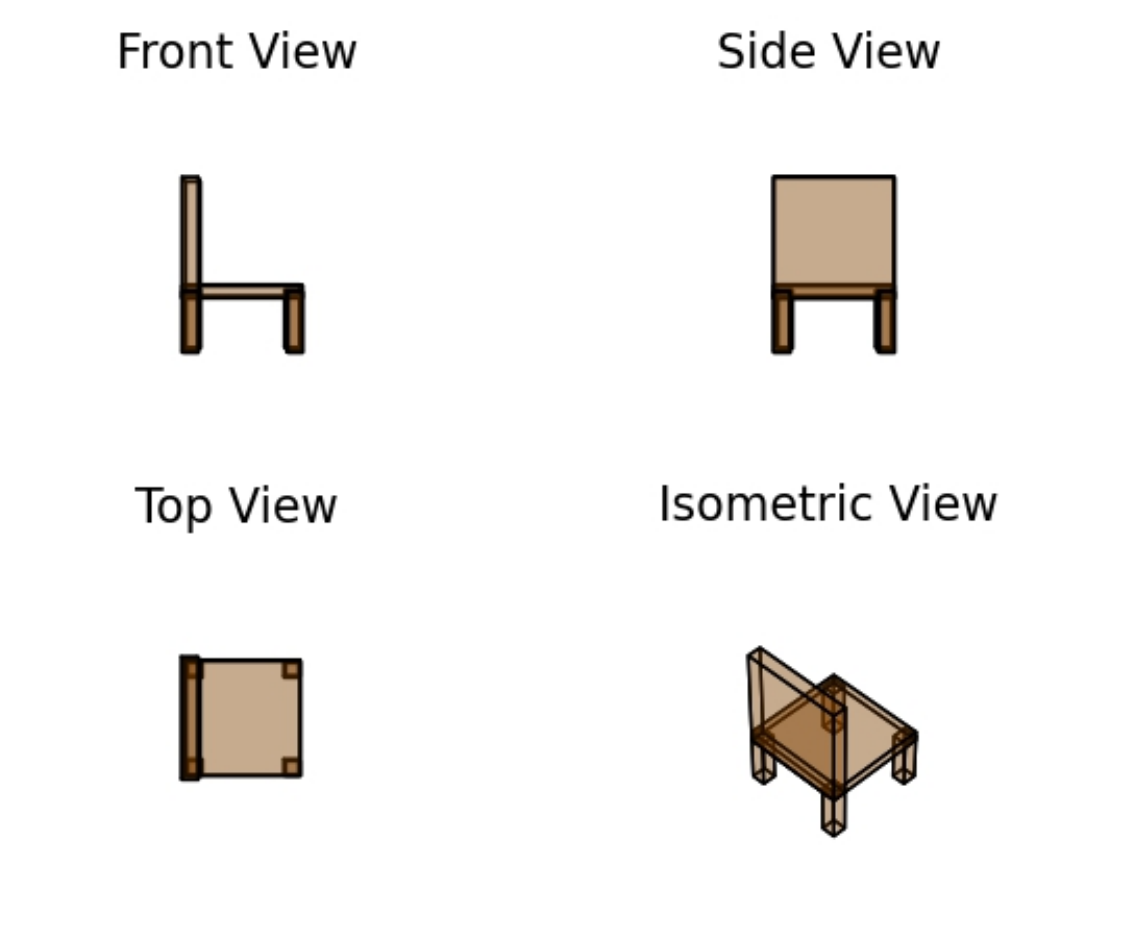}
        \caption{Chair(4)}
        \label{fig:chair4}
    \end{subfigure}
    \hfill
    \begin{subfigure}[b]{0.18\textwidth}
        \centering
        \includegraphics[width=\textwidth]{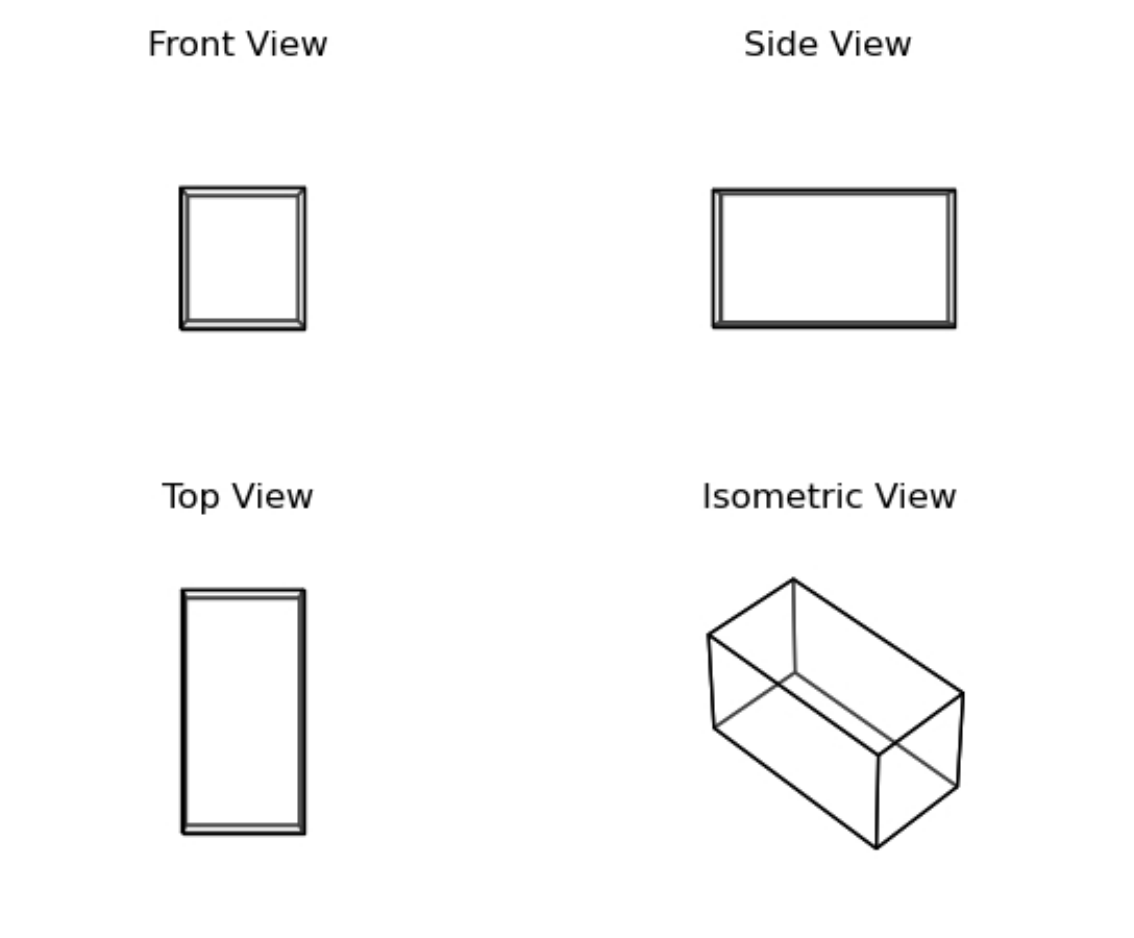}
        \caption{Table(3.5)}
        \label{fig:table3}
    \end{subfigure}
    \hfill
    \begin{subfigure}[b]{0.18\textwidth}
        \centering
        \includegraphics[width=\textwidth]{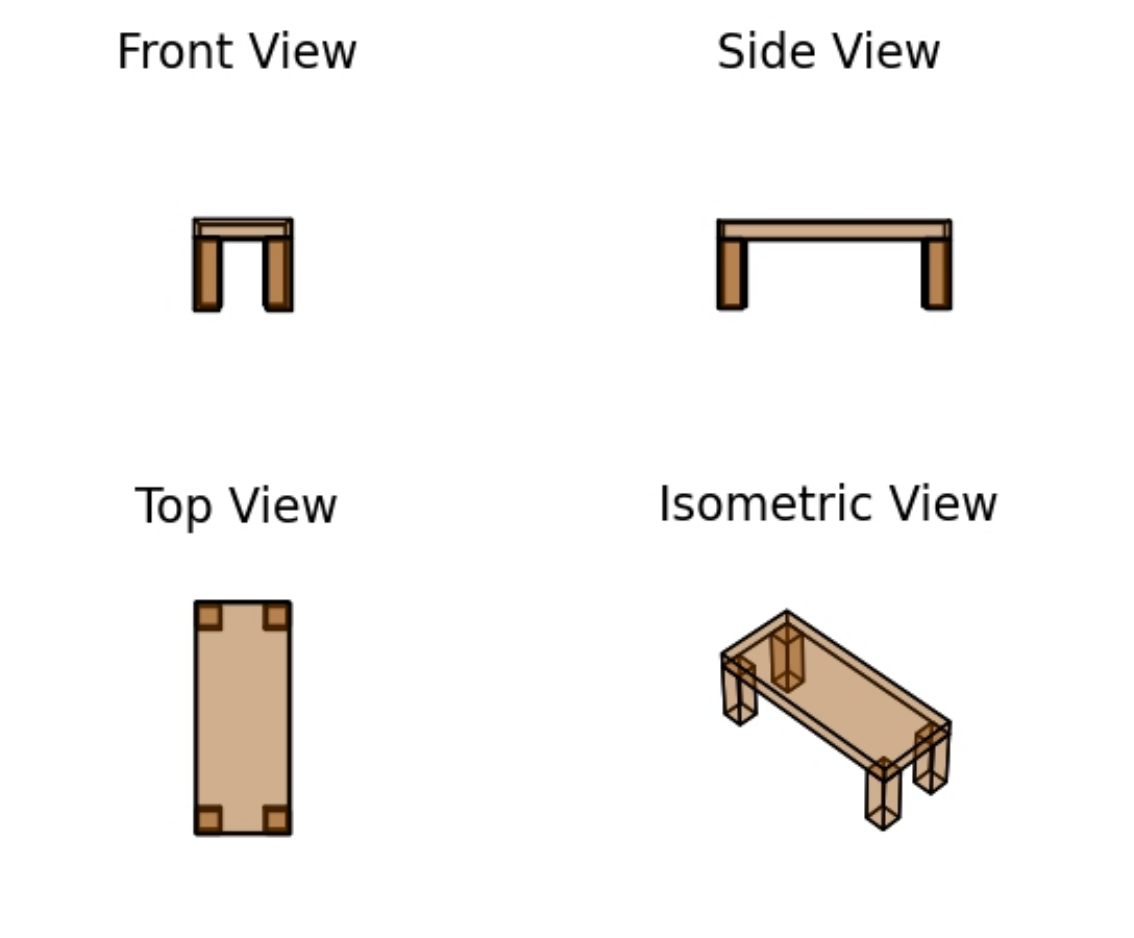}
        \caption{Tabel(4)}
        \label{fig:tabel4}
    \end{subfigure}
    \begin{subfigure}[b]{0.18\textwidth}
        \centering
        \includegraphics[width=\textwidth]{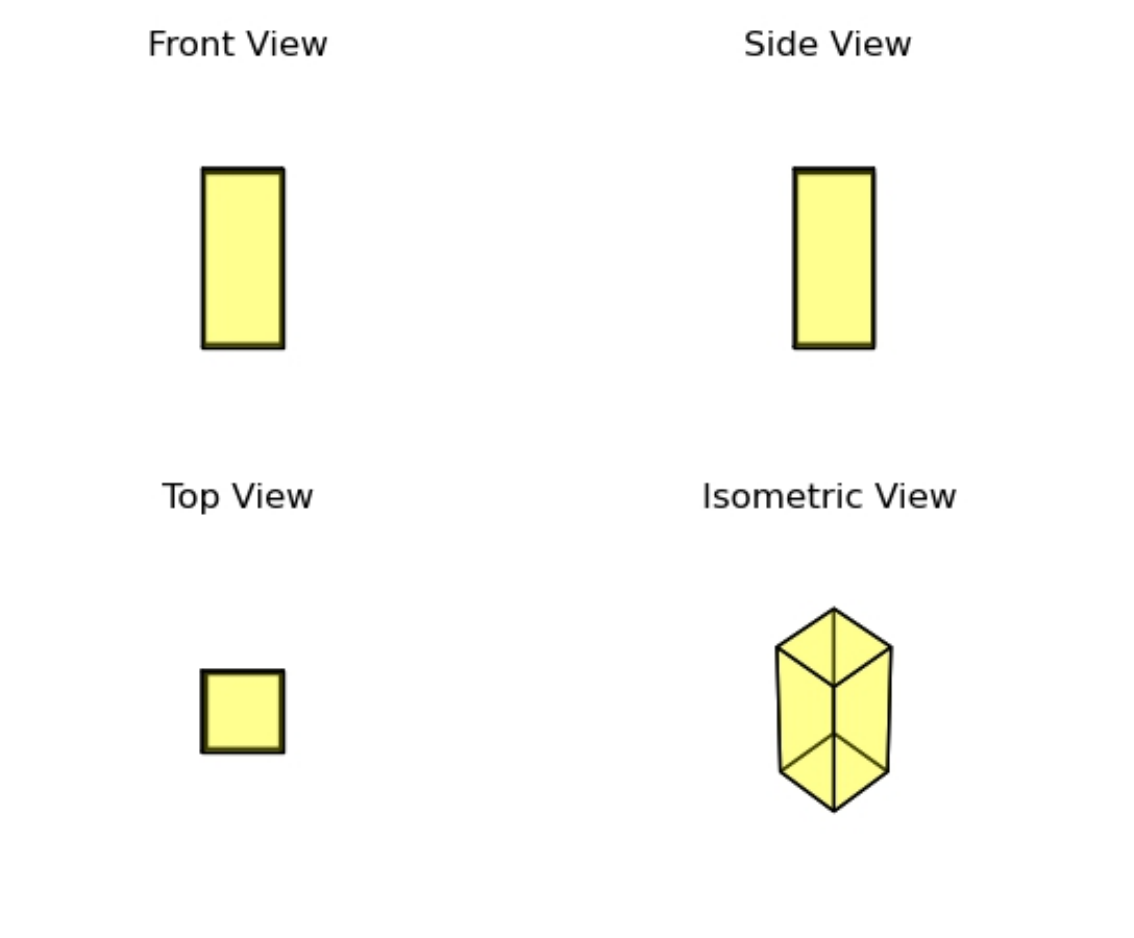}
        \caption{Lamp(3.5)}
        \label{fig:lamp3}
    \end{subfigure}
    \hfill
    \begin{subfigure}[b]{0.18\textwidth}
        \centering
        \includegraphics[width=\textwidth]{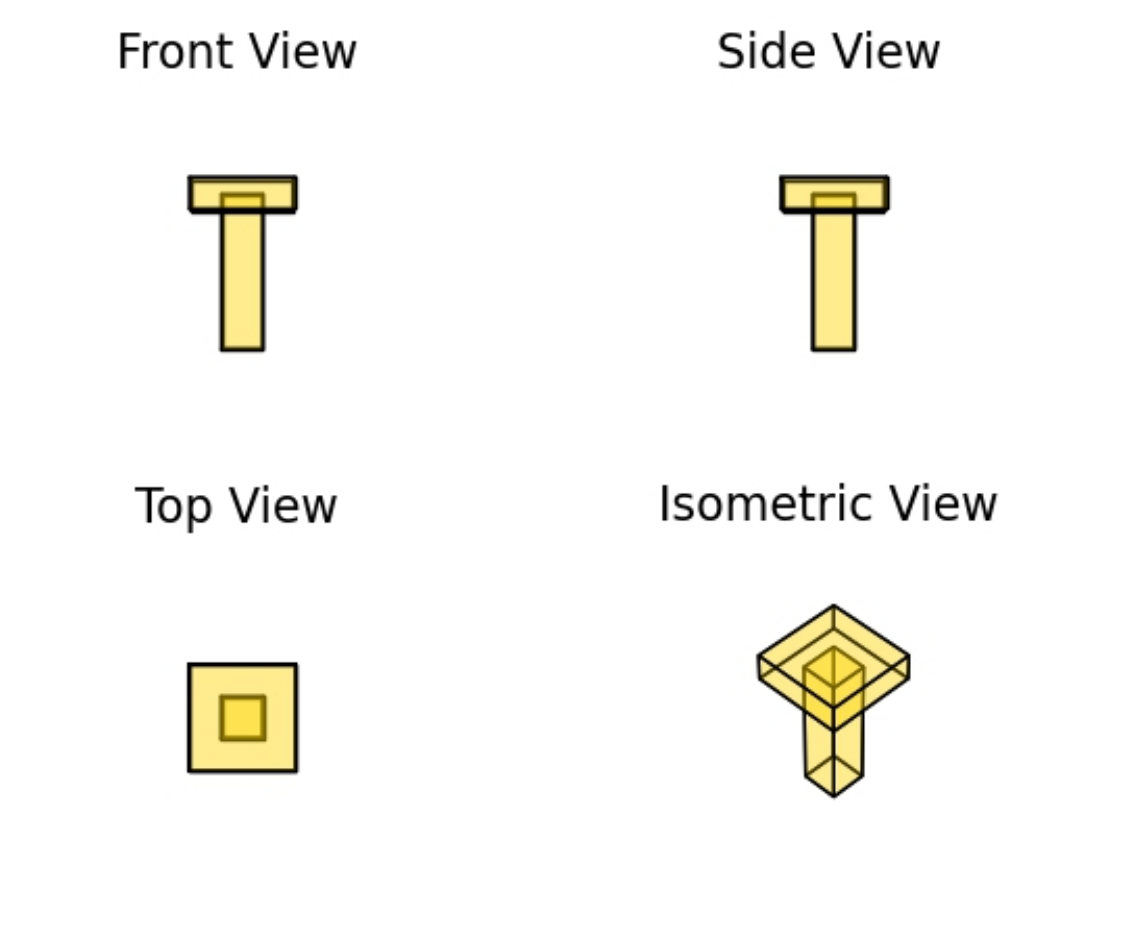}
        \caption{Lamp(4)}
        \label{fig:lamp4}
    \end{subfigure}
    \caption{GPT-4 produces complex structures and details and achieves better semantic alignment than GPT-3.5.}
    \label{fig:baseline_all_figures_obj}
\end{figure}

\subsubsection{Scenery Generation}
\begin{figure}[H]
    \centering
    \begin{subfigure}[b]{0.18\textwidth}
        \centering
        \includegraphics[width=\textwidth]{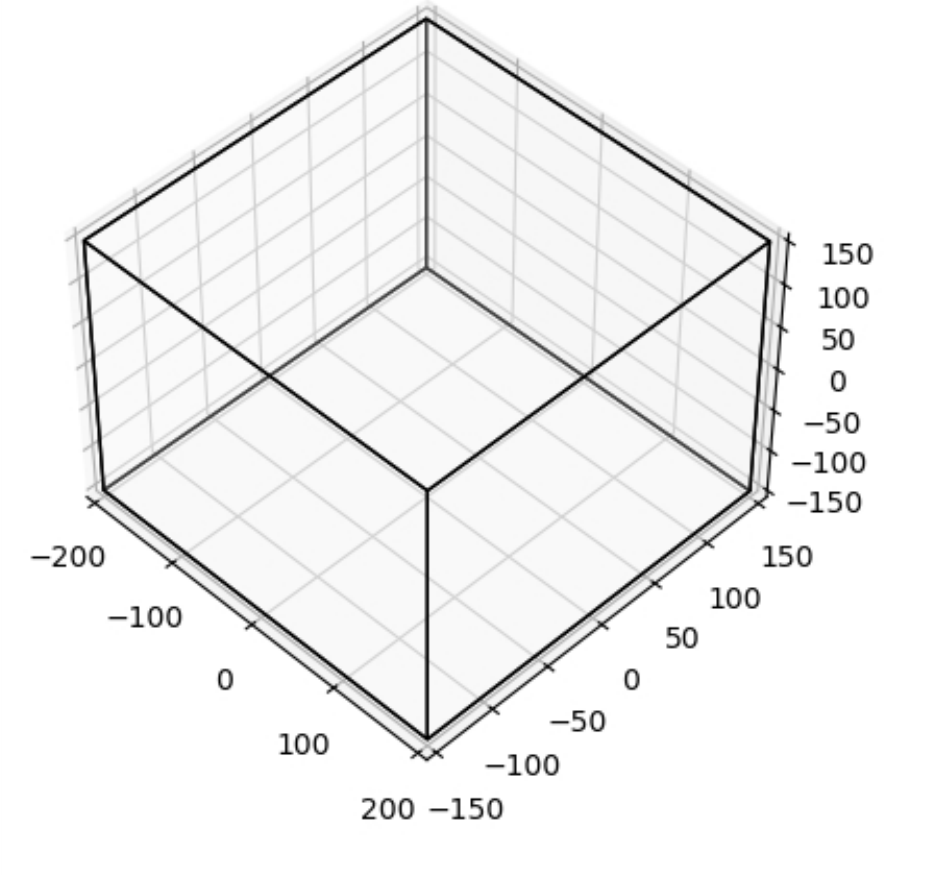}
        \caption{Bedroom(3.5)}
        \label{fig:baseline_bedroom3}
    \end{subfigure}
    \hfill
    \begin{subfigure}[b]{0.18\textwidth}
        \centering
        \includegraphics[width=\textwidth]{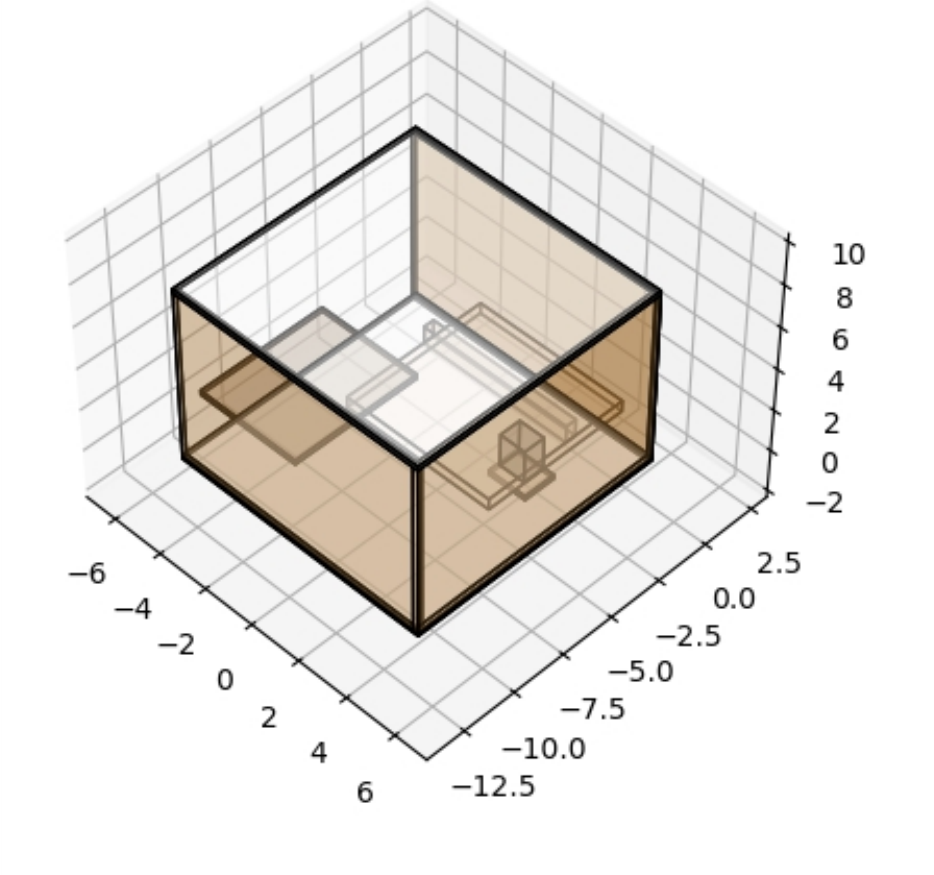}
        \caption{Bedroom(4)}
        \label{fig:baseline_bedroom4}
    \end{subfigure}
    \hfill
    \begin{subfigure}[b]{0.18\textwidth}
        \centering
        \includegraphics[width=\textwidth]{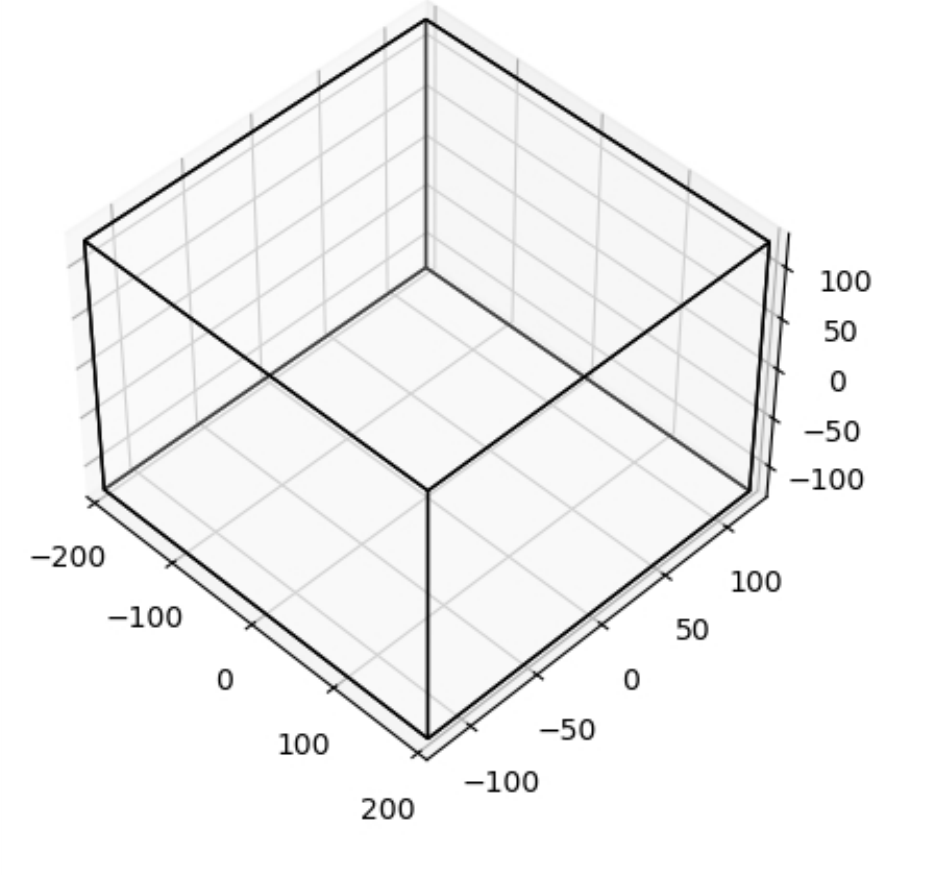}
        \caption{Kitchen(3.5)}
        \label{fig:baseline_kitchen3}
    \end{subfigure}
    \hfill
    \begin{subfigure}[b]{0.18\textwidth}
        \centering
        \includegraphics[width=\textwidth]{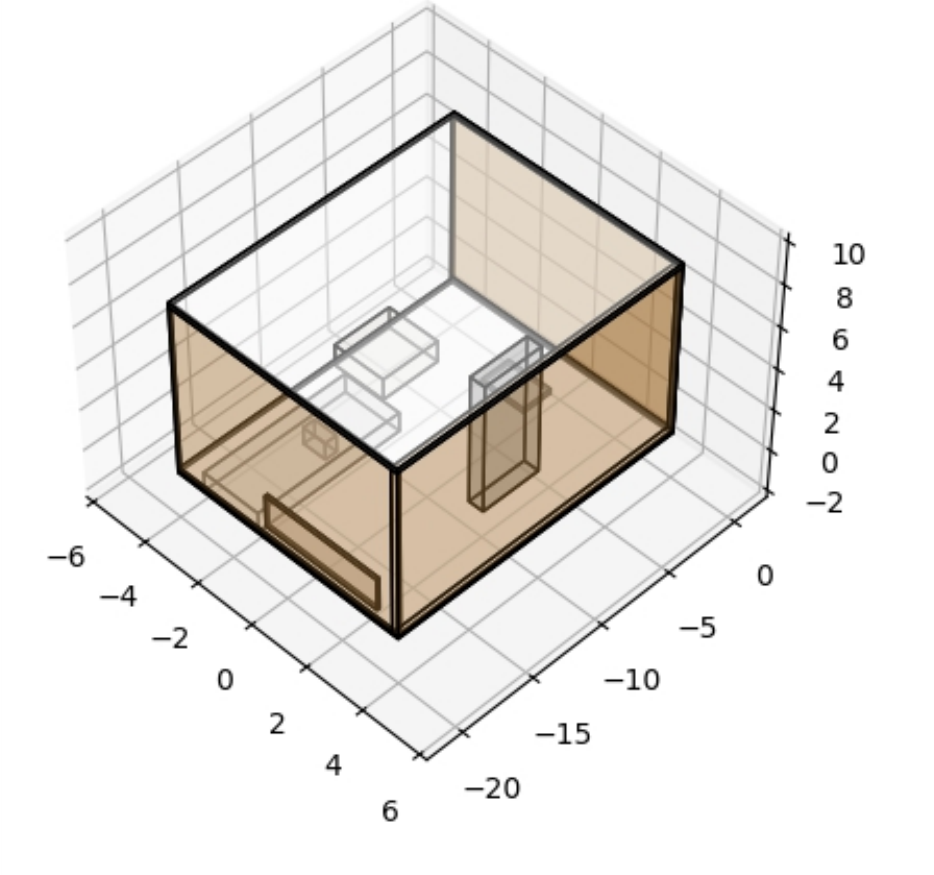}
        \caption{Kitchen(4)}
        \label{fig:baseline_kitchen4}
    \end{subfigure}
    \begin{subfigure}[b]{0.18\textwidth}
        \centering
        \includegraphics[width=\textwidth]{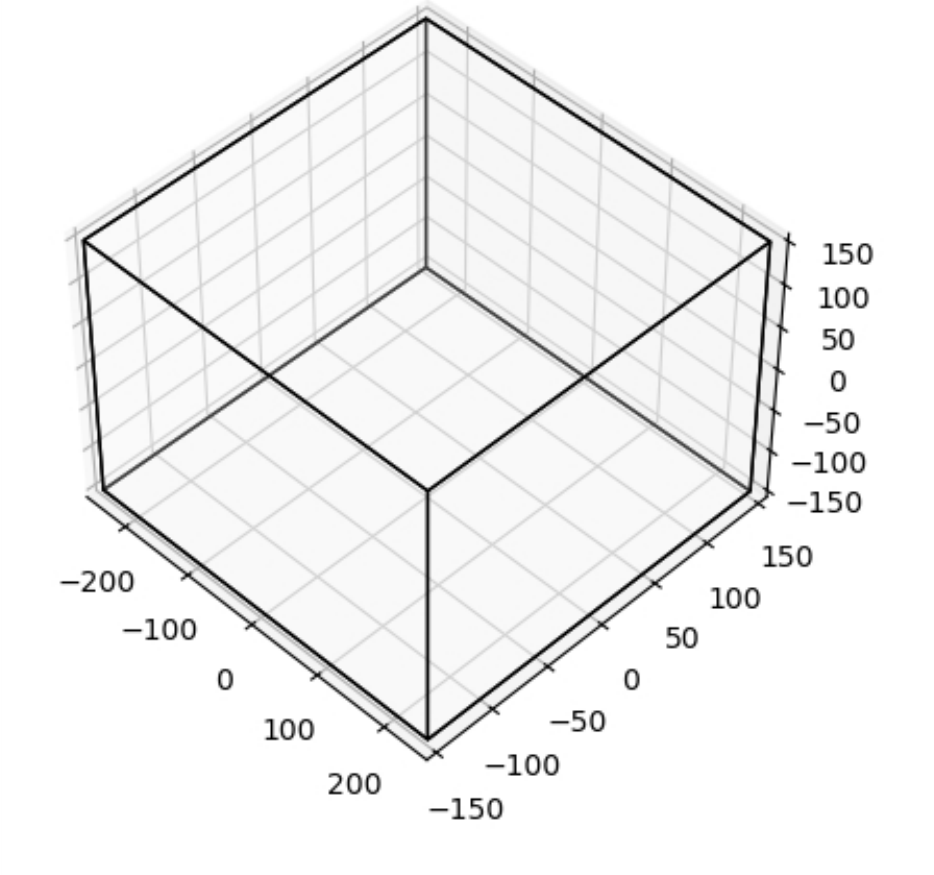}
        \caption{Living(3.5)}
        \label{fig:baseline_living3}
    \end{subfigure}
    \hfill
    \begin{subfigure}[b]{0.18\textwidth}
        \centering
        \includegraphics[width=\textwidth]{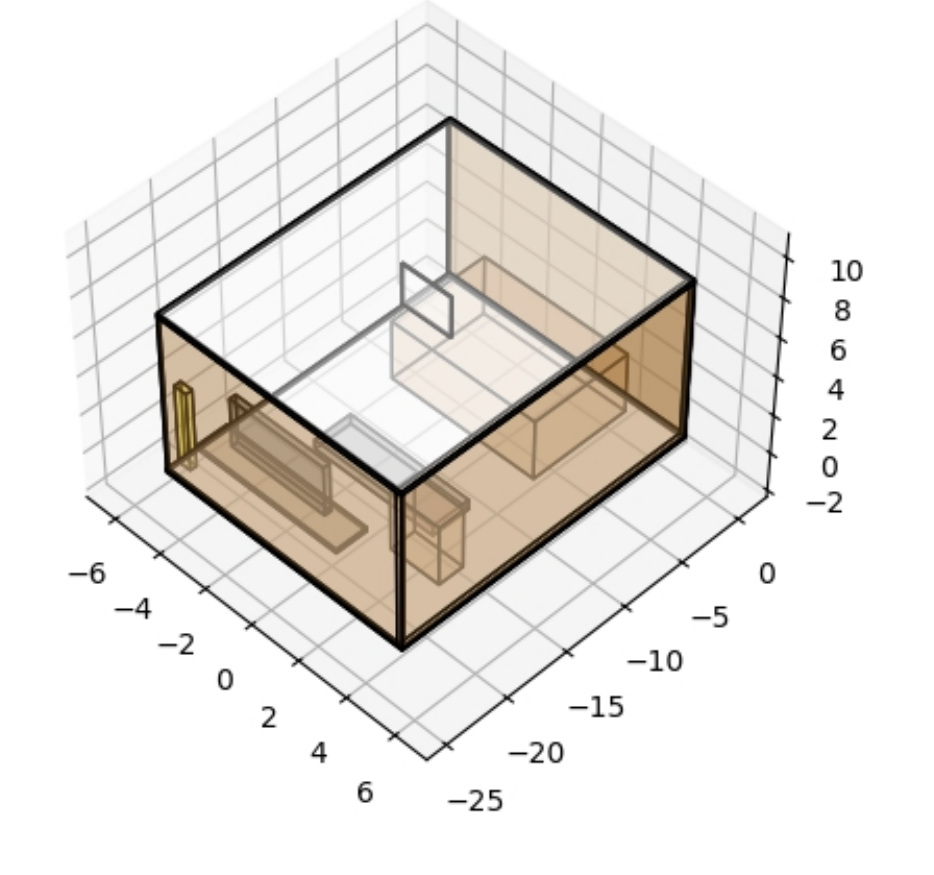}
        \caption{Living(4)}
        \label{fig:baseline_living4}
    \end{subfigure}
    \caption{GPT-4 shows better spatial comprehension and multi-object scene generation than GPT-3.5, but still uses simple blocks with limited detail.}
    \label{fig:baseline_all_figures_scenery}
\end{figure}
\subsection{Analysis And Comparison}
\textbf{Metric:}We choose CLIP \citep{radford2021learning}to calculate the similarity between the generated object and scene images and text, in order to evaluate the alignment between the text and the generated content. In addition, during the experimental process, there is often a large amount of overlap or object isolation in the generated failed scene images. Therefore, for the scene, we additionally introduced overlap score and isolation score, corresponding to the proportion of overlapping volume to the total volume of all objects and the proportion of isolated blocks to the total block, respectively.
\begin{figure}[H]
    \centering
    \includegraphics[width=1\linewidth]{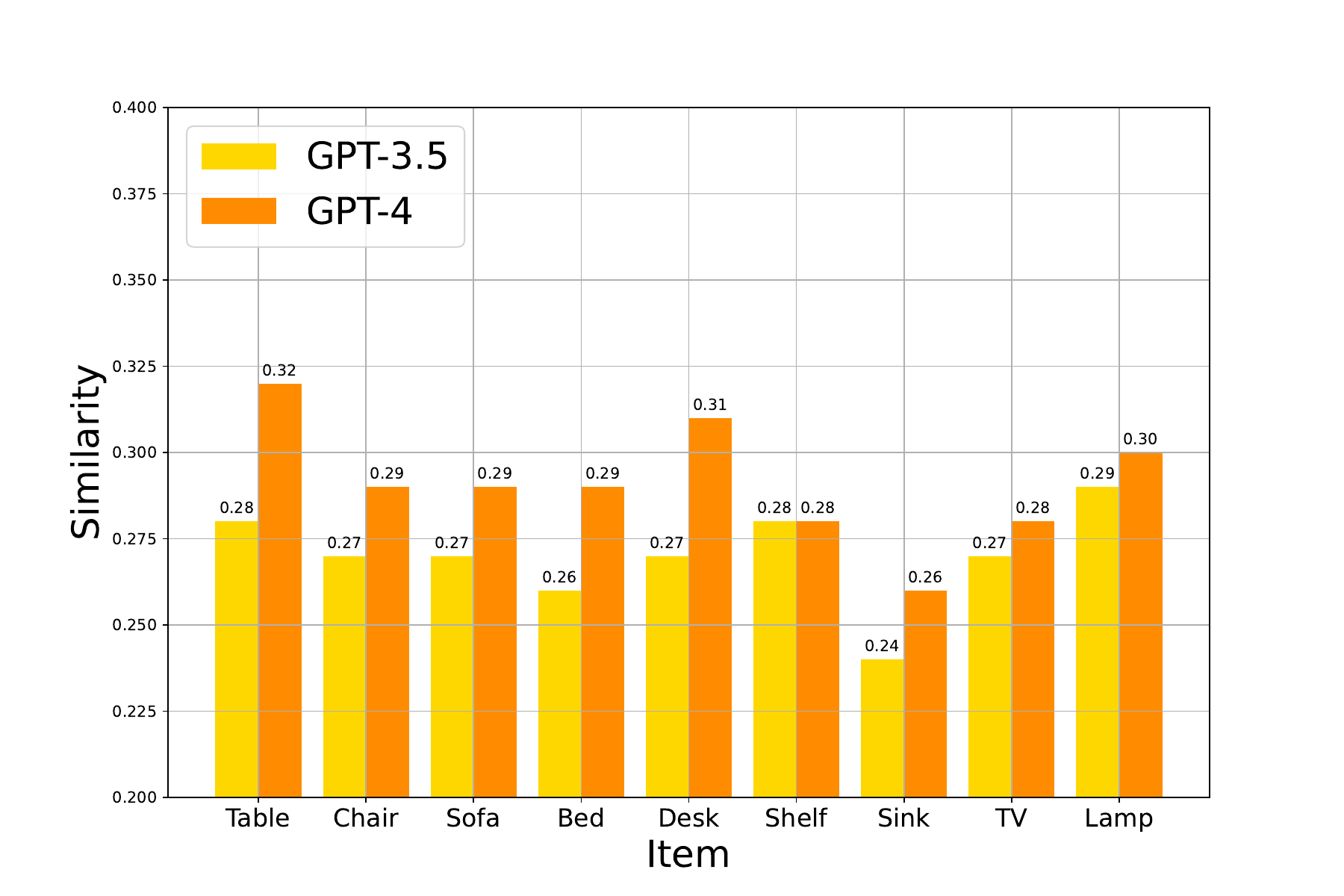}
    \caption{In object level generation tasks, the clip index of agents based on GPT-4 is generally better than  ones based on GPT-3.5.}
    \label{fig:item_similarity_comparison}
\end{figure}

\begin{figure}[H]
    \centering
    \includegraphics[width=1\linewidth]{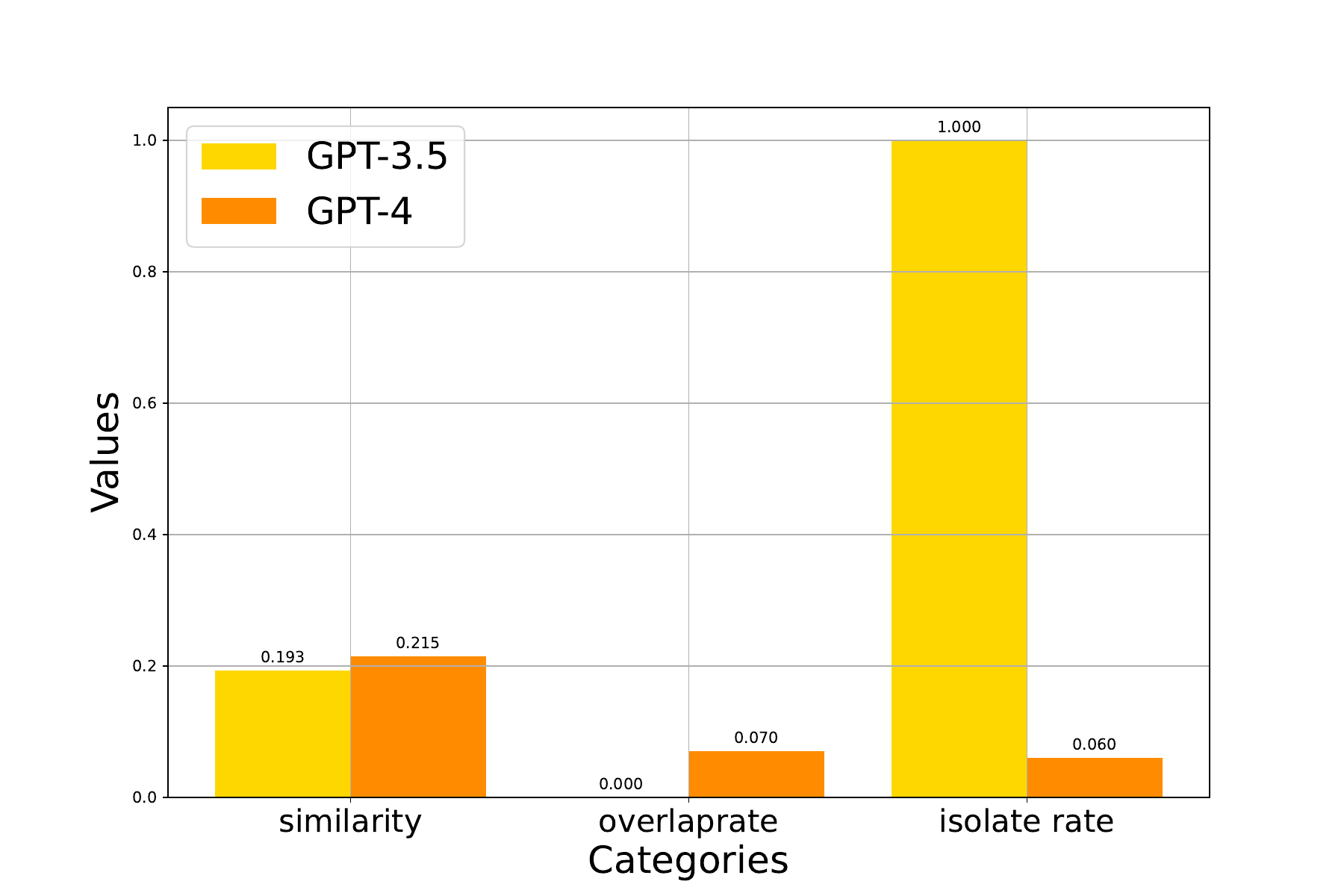}
    \caption{In the scenario level generation task, the clip index of GPT-4 group is \textbf{10.1\%} higher than that of GPT-3.5 group, and its isolation rate is much better than that of GPT-3.5 group.}
    \label{fig:comparison_plot}
\end{figure}

\subsection{Framework Impact}

\textbf{Baseline Methods:}The baseline we have chosen is a single agent without designed agents or graph driven methods, which showed in Figure \ref{fig:baseline_all_figures_scenery}. The base model of each agent is gpt-4-0125 preview with default temperature.

\subsubsection{Ablation Study}
\begin{figure}[H]
    \centering
    \begin{subfigure}[b]{0.18\textwidth}
        \centering
        \includegraphics[width=\textwidth]{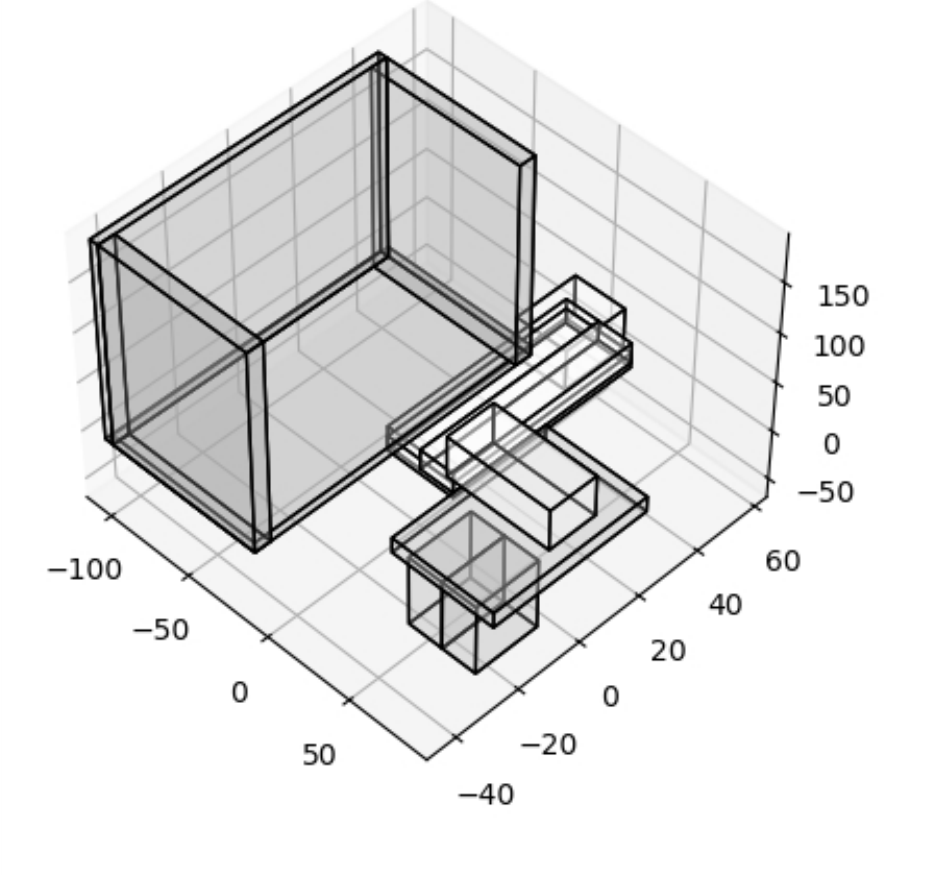}
        \caption{Bathroom(ablation)}
        \label{fig:chain_bathroom}
    \end{subfigure}
    \hfill
    \begin{subfigure}[b]{0.18\textwidth}
        \centering
        \includegraphics[width=\textwidth]{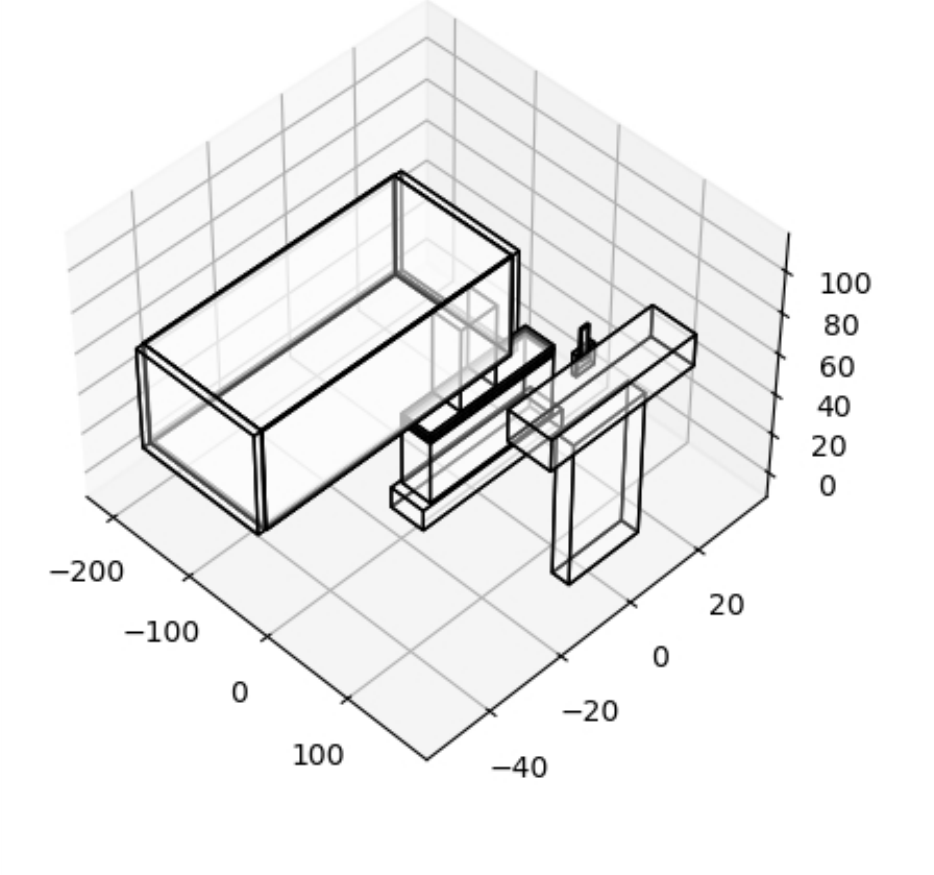}
        \caption{Bathroom}
        \label{fig:final_bathroom}
    \end{subfigure}
    \hfill
    \begin{subfigure}[b]{0.18\textwidth}
        \centering
        \includegraphics[width=\textwidth]{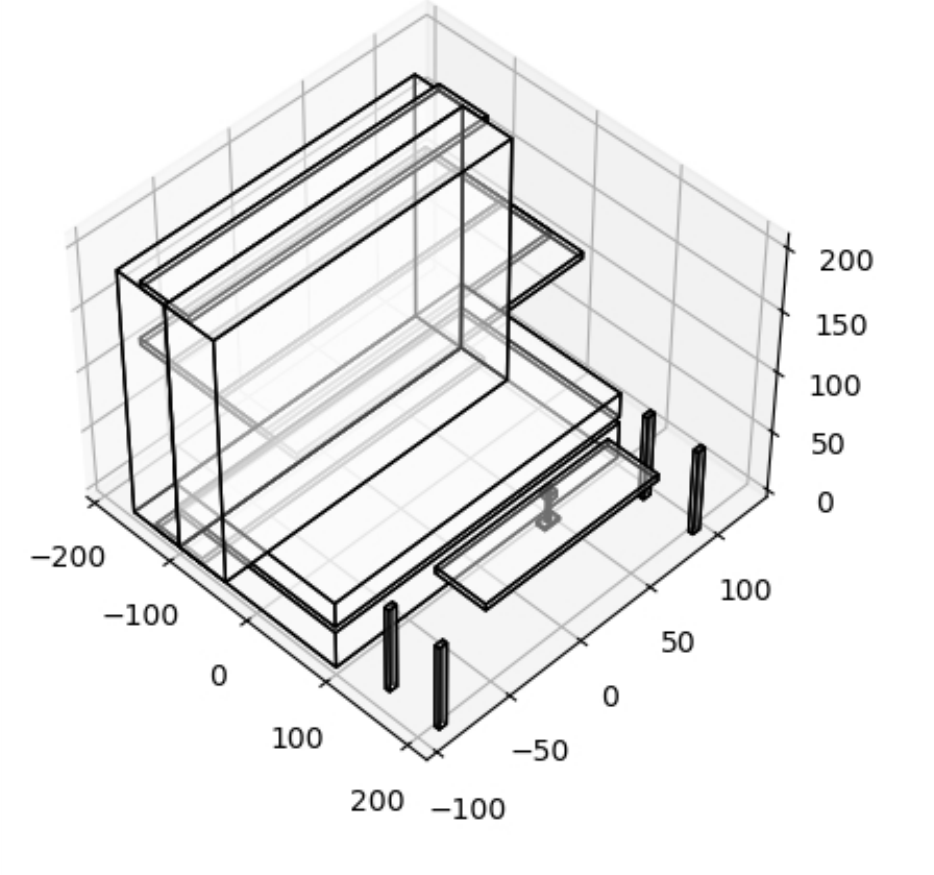}
        \caption{Bedroom (ablation)}
        \label{fig:chain_bedroom}
    \end{subfigure}
    \hfill
    \begin{subfigure}[b]{0.18\textwidth}
        \centering
        \includegraphics[width=\textwidth]{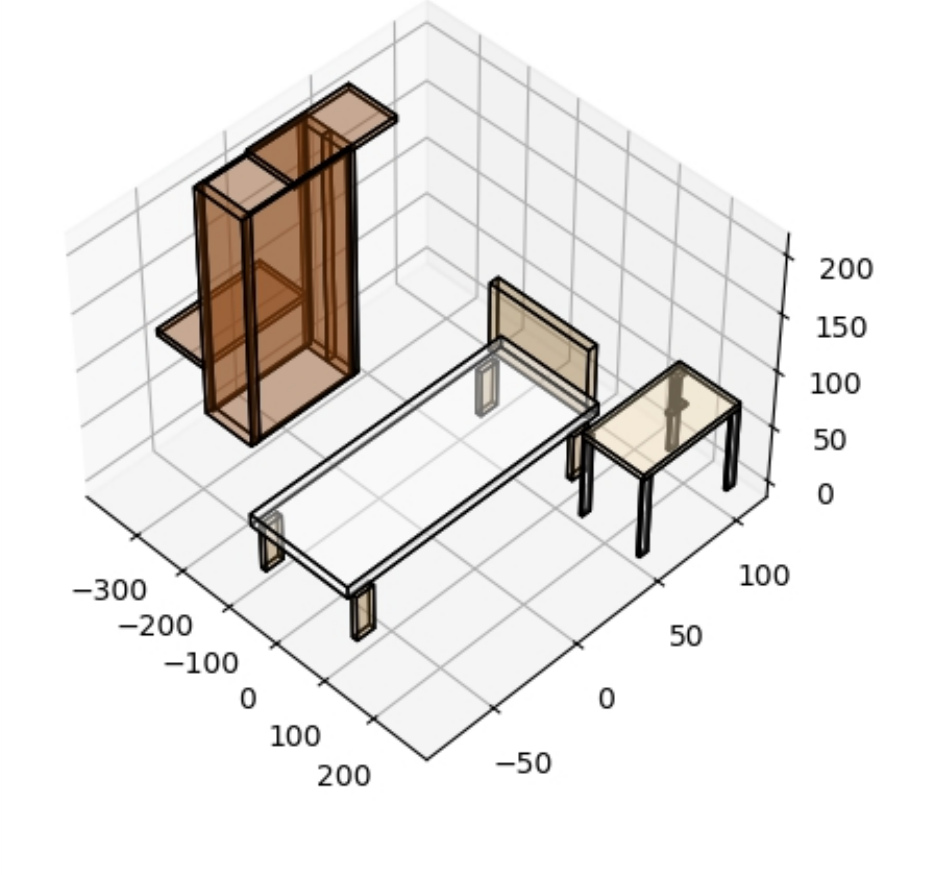}
        \caption{Bedroom}
        \label{fig:final_bedroom}
    \end{subfigure}
    \hfill
    \begin{subfigure}[b]{0.18\textwidth}
        \centering
        \includegraphics[width=\textwidth]{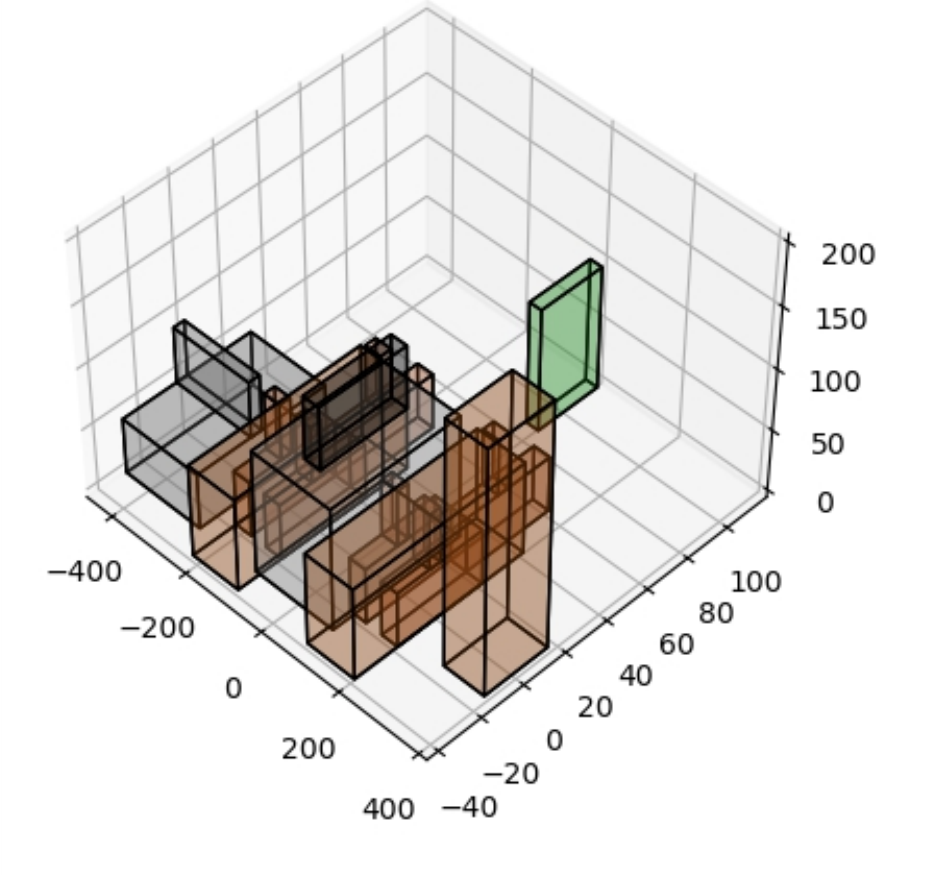}
        \caption{Cafe (ablation)}
        \label{fig:chain_cafe}
    \end{subfigure}
    \hfill
    \begin{subfigure}[b]{0.18\textwidth}
        \centering
        \includegraphics[width=\textwidth]{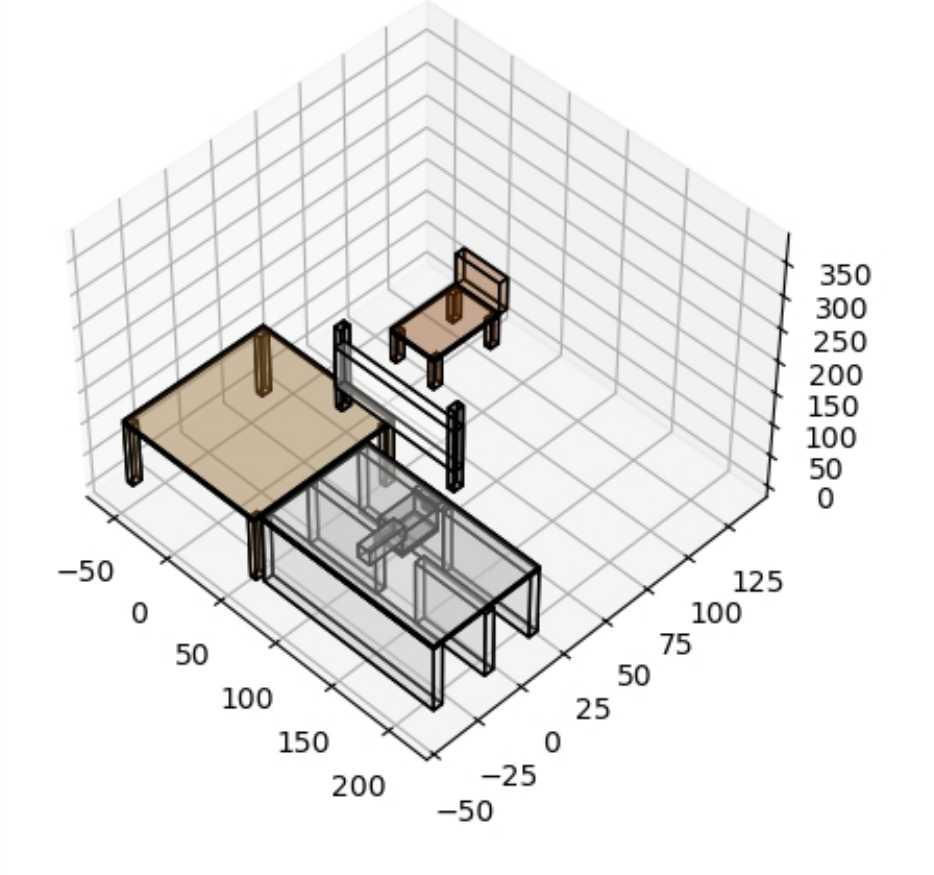}
        \caption{Cafe}
        \label{fig:final_cafe}
    \end{subfigure}
    \hfill
    \begin{subfigure}[b]{0.18\textwidth}
        \centering
        \includegraphics[width=\textwidth]{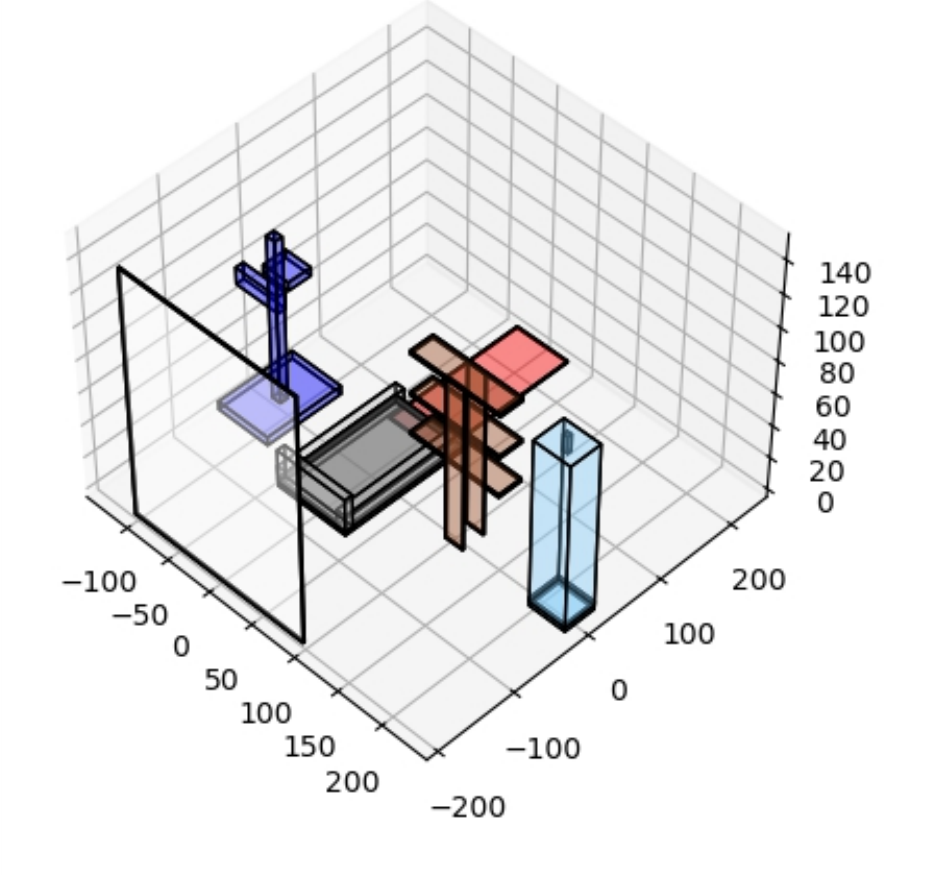}
        \caption{Gym (ablation)}
        \label{fig:chain_gym}
    \end{subfigure}
    \hfill
    \begin{subfigure}[b]{0.18\textwidth}
        \centering
        \includegraphics[width=\textwidth]{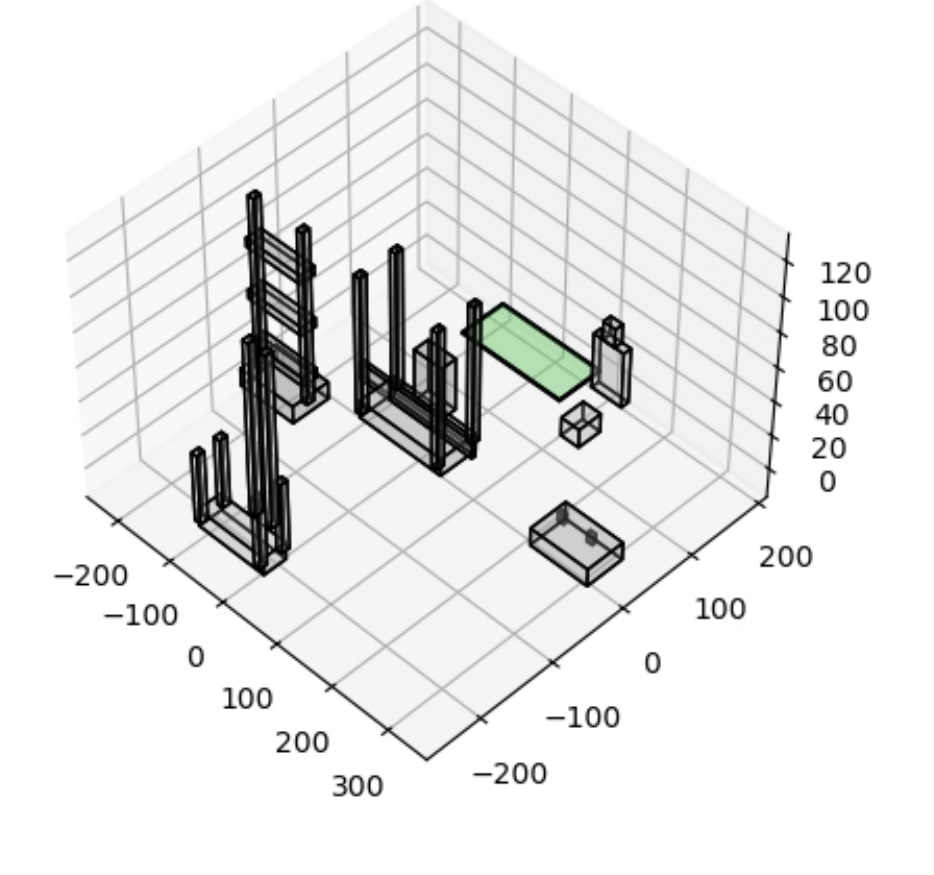}
        \caption{Gym}
        \label{fig:final_gym}
    \end{subfigure}
    \hfill
    \begin{subfigure}[b]{0.18\textwidth}
        \centering
        \includegraphics[width=\textwidth]{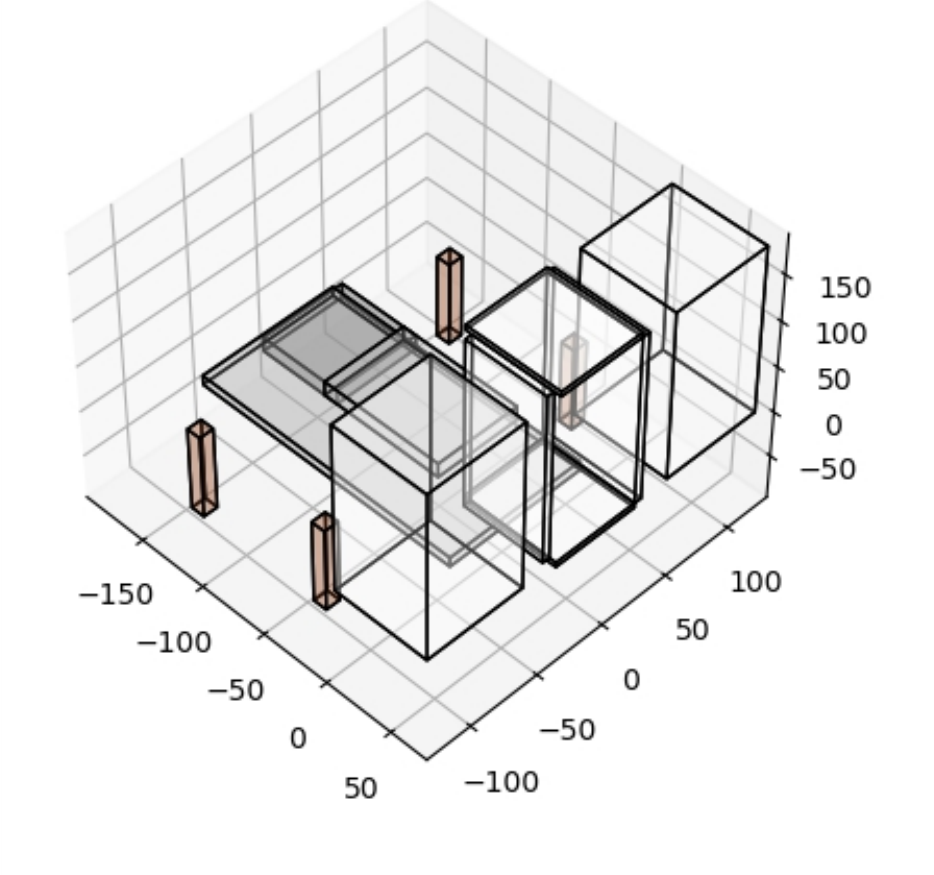}
        \caption{Kitchen (ablation)}
        \label{fig:chain_kitchen}
    \end{subfigure}
    \hfill
    \begin{subfigure}[b]{0.18\textwidth}
        \centering
        \includegraphics[width=\textwidth]{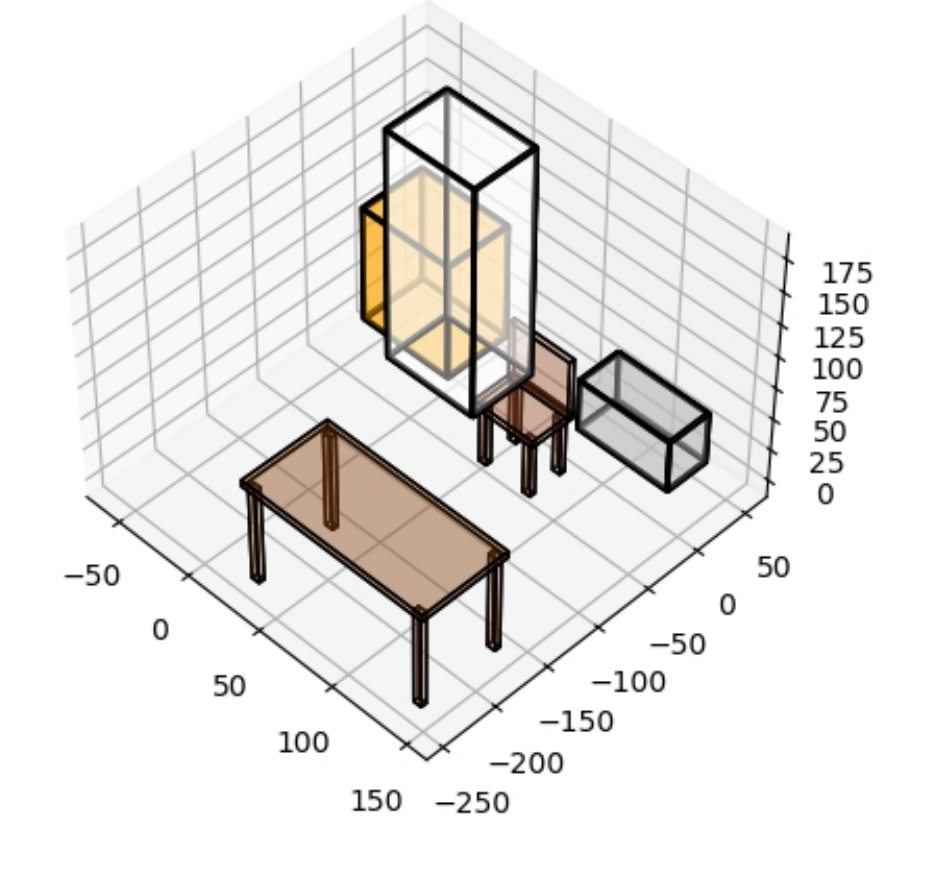}
        \caption{Kitchen}
        \label{fig:final_kitchen}
    \end{subfigure}
    \hfill
    \begin{subfigure}[b]{0.18\textwidth}
        \centering
        \includegraphics[width=\textwidth]{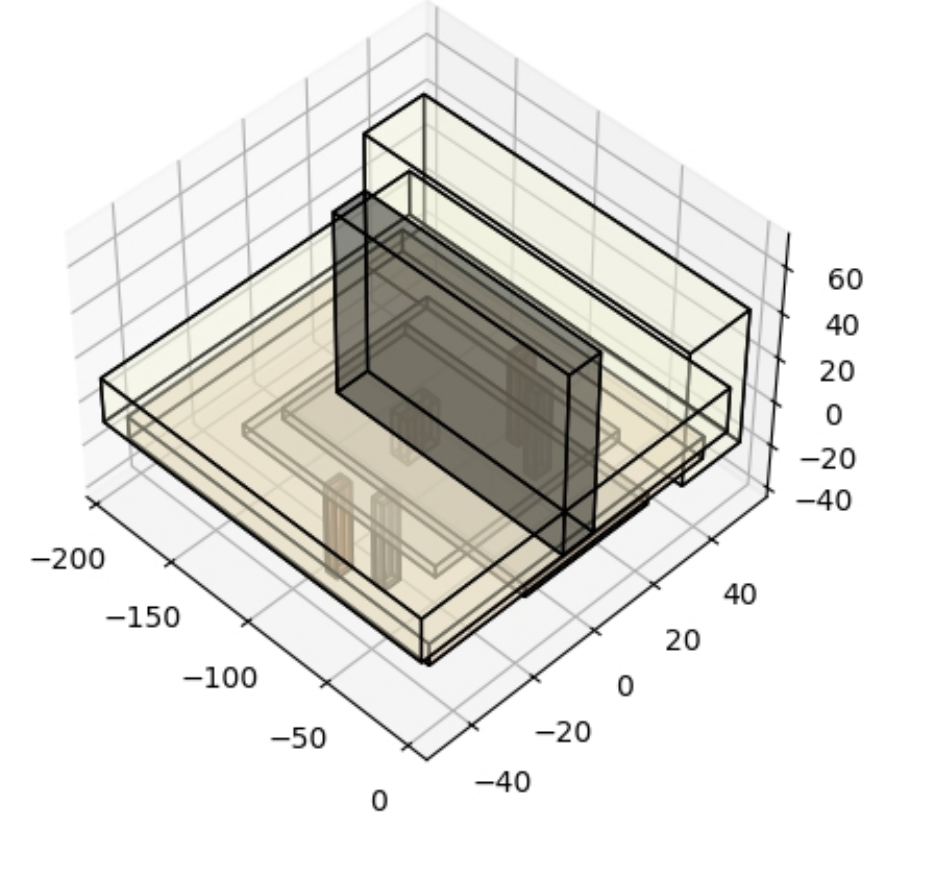}
        \caption{Living (ablation)}
        \label{fig:chain_livingroom}
    \end{subfigure}
    \hfill
    \begin{subfigure}[b]{0.18\textwidth}
        \centering
        \includegraphics[width=\textwidth]{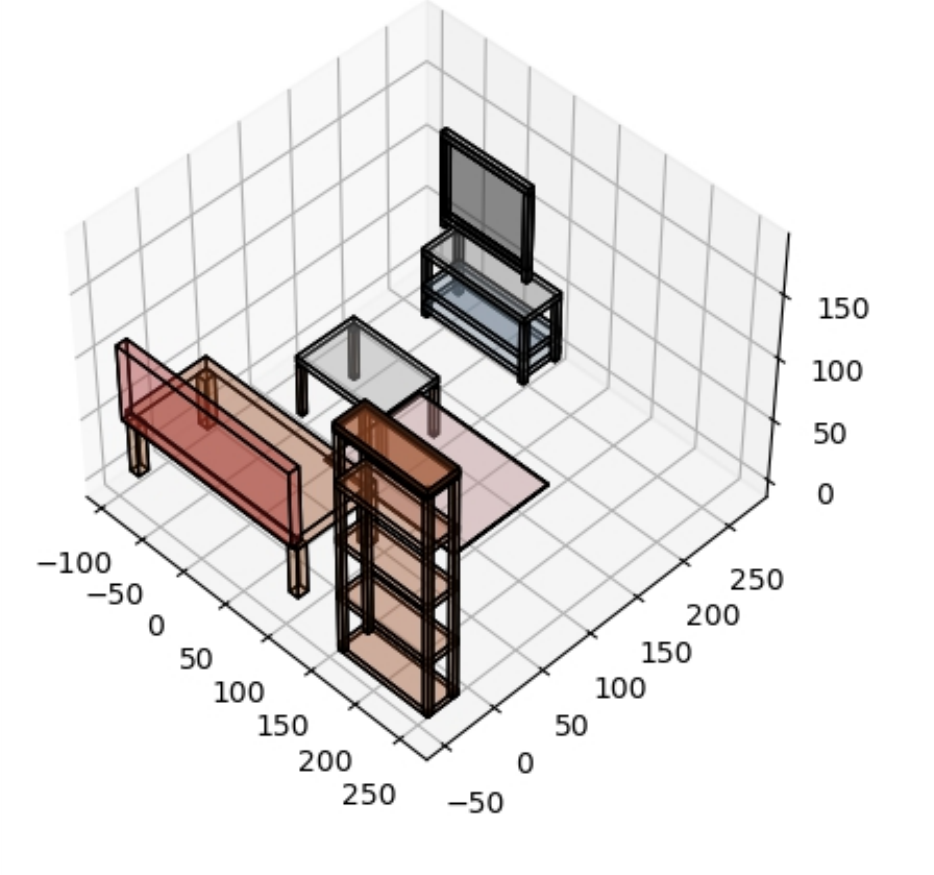}
        \caption{Living}
        \label{fig:final_livingroom}
    \end{subfigure}
    \caption{The ablation group showed detailed structures, but lacked reasonable spatial planning  The non-ablated group can not only represent details of objects but also have a reasonable plan for the placement of objects.
    }
    \label{fig:figure comparsion}
\end{figure}

In the ablation group experiment, we eliminated the interaction process between the graphical database and the workflow, while retaining the workflow of multi-agent collaboration. The non ablated group completely retained the graph reasoning framework.

\subsection{Analysis And Comparison}

The schematic diagram illustrates the performance of LLM scene graph generation in three modes. Images produced by the baseline method neglect object details but exhibit some overall spatial planning capability. The ablation group attempts to emphasize object details but lacks spatial planning, leading to overcrowded scenes. The non-ablated group excels in both object details and proper object placement.
\begin{figure}[H]
    \centering
    \includegraphics[width=1\linewidth]{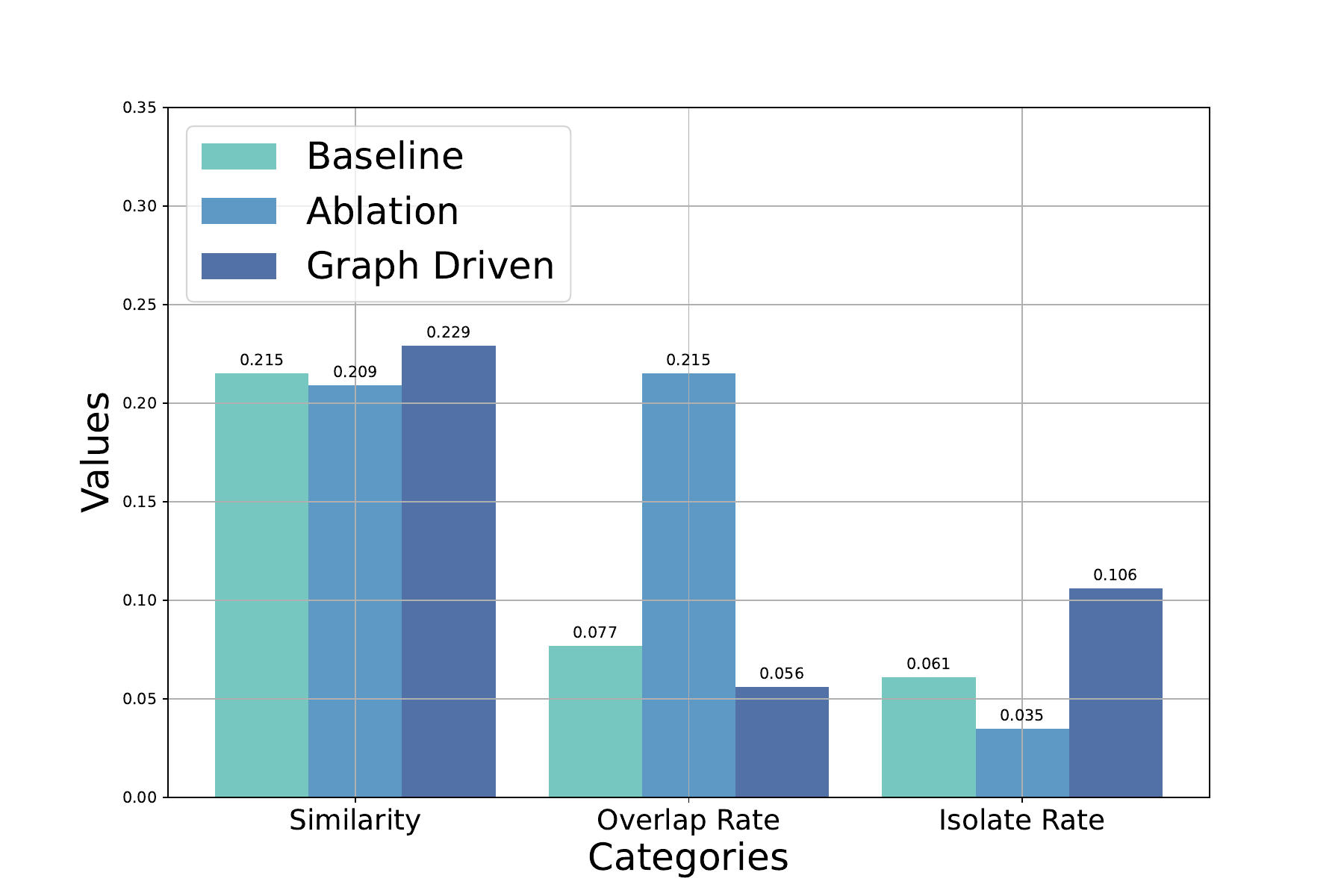}
    \caption{Comparison of metrics across different work modes indicates the following information: using graph driven workflows improves the similarity between images and text, with a decrease in spatial overlap rate but an increase in isolation rate}
    \label{fig:comparison}
\end{figure}
According to the above Figure \ref{fig:comparison}, we found that in terms of clip similarity, the graph driven group performed better than both the baseline and ablation groups, and was generally better than both in a single task, with mean values \textbf{6.3\% }and \textbf{8.7\%} higher than the baseline and ablation groups, respectively. In terms of object overlap rate, it is lower than both, but in terms of isolation rate, it is higher than both.

\section{Conclusion And Limitation}
Our research provides an intuitive demonstration of the spatial understanding capabilities of LLMs and quantitatively evaluates the spatial comprehension of two distinct models. Additionally, we enhance the geometric understanding and spatial reasoning abilities of LLMs in complex physical environments by implementing well-defined geometric conventions and a graph-driven framework.

This study is conducted using a custom-developed sandbox platform, designed to present the spatial concepts understood by LLMs in a more intuitive and flexible manner. However, due to resource constraints, we are unable to test higher-performing models, which limits our ability to fully showcase the framework's potential in improving the spatial understanding of LLMs.

\bibliography{reference}

\appendix
\section{Geometric conventions}
\subsection{Geometric Center Relationship Constrain}
\begin{enumerate}
\item \textbf{concentric}: Concentric.
        \begin{itemize}
            \item Calculation: The Euclidean distance between the centers of the two objects.
        \end{itemize}
\end{enumerate}
\subsection{Axle Relationship Constraint}
\subsubsection{Align Relationship}
    \begin{enumerate}
    \item \textbf{x aligned}: X-aligned.
        \begin{itemize}
            \item Calculation: The alignment error in the x direction between the two objects.
        \end{itemize}
        
    \item \textbf{y aligned}: Y-aligned.
        \begin{itemize}
            \item Calculation: The alignment error in the y direction between the two objects.
        \end{itemize}
        
    \item \textbf{z aligned}: Z-aligned.
        \begin{itemize}
            \item Calculation: The alignment error in the z direction between the two objects.
        \end{itemize}
    \end{enumerate}
\subsubsection{Half Side Relationship}
\begin{enumerate}
 \item \textbf{left\ half}: Left half.
        \begin{itemize}
            \item Determine if the \texttt{ref center} object is in the left half of the \texttt{mov center} object.
        \end{itemize}
        
    \item \textbf{right\ half}: Right half.
        \begin{itemize}
            \item Determine if the \texttt{ref center} object is in the right half of the \texttt{mov center} object.
        \end{itemize}
        
    \item \textbf{upper\ half}: Upper half.
        \begin{itemize}
            \item Determine if the \texttt{ref center} object is in the upper half of the \texttt{mov center} object.
        \end{itemize}
        
    \item \textbf{lower\ half}: Lower half.
        \begin{itemize}
            \item Determine if the \texttt{ref center} object is in the lower half of the \texttt{mov center} object.
        \end{itemize}
        
    \item \textbf{front\ half}: Front half.
        \begin{itemize}
            \item Determine if the \texttt{ref center} object is in the front half of the \texttt{mov center} object.
        \end{itemize}
        
    \item \textbf{back\ half}: Back half.
        \begin{itemize}
            \item Determine if the \texttt{ref center} object is in the back half of the \texttt{mov center} object.
        \end{itemize}
    \end{enumerate}
\subsection{Surface Relationship Constraint}
\subsubsection{Relative Positioning Relationship}
\begin{enumerate}
    \item \textbf{left}: \texttt{mov center} object is to the left of the \texttt{ref center} object.
        \begin{itemize}
            \item Calculation: The distance between the left edge of the \texttt{ref center} object and the right edge of the \texttt{mov center} object minus the given \texttt{distance}.
        \end{itemize}
        
    \item \textbf{right}: \texttt{mov center} object is to the right of the \texttt{ref center} object.
        \begin{itemize}
            \item Calculation: The distance between the left edge of the \texttt{mov center} object and the right edge of the \texttt{ref center} object minus the given \texttt{distance}.
        \end{itemize}
        
    \item \textbf{above}: \texttt{mov center} object is above the \texttt{ref center} object.
        \begin{itemize}
            \item Calculation: The distance between the bottom edge of the \texttt{mov center} object and the top edge of the \texttt{ref center} object minus the given \texttt{distance}.
        \end{itemize}
        
    \item \textbf{below}: \texttt{mov center} object is below the \texttt{ref center} object.
        \begin{itemize}
            \item Calculation: The distance between the bottom edge of the \texttt{ref center} object and the top edge of the \texttt{mov center} object minus the given \texttt{distance}.
        \end{itemize}
        
    \item \textbf{front}: \texttt{mov center} object is in front of the \texttt{ref center} object.
        \begin{itemize}
            \item Calculation: The distance between the back edge of the \texttt{mov center} object and the front edge of the \texttt{ref center} object minus the given \texttt{distance}.
        \end{itemize}
        
    \item \textbf{back}: \texttt{mov center} object is behind the \texttt{ref center} object.
        \begin{itemize}
            \item Calculation: The distance between the back edge of the \texttt{ref center} object and the front edge of the \texttt{mov center} object minus the given \texttt{distance}.
        \end{itemize}
    \end{enumerate}
    
    \subsubsection{Coplanar Relationship Constrain}
    \begin{enumerate}
    \item \textbf{coplanar top}: Coplanar on top.
        \begin{itemize}
            \item Determine if the top edges of the two objects are coplanar.
        \end{itemize}
        
    \item \textbf{coplanar bottom}: Coplanar on the bottom.
        \begin{itemize}
            \item Determine if the bottom edges of the two objects are coplanar.
        \end{itemize}
        
    \item \textbf{coplanar left}: Coplanar on the left.
        \begin{itemize}
            \item Determine if the left edges of the two objects are coplanar.
        \end{itemize}
        
    \item \textbf{coplanar right}: Coplanar on the right.
        \begin{itemize}
            \item Determine if the right edges of the two objects are coplanar.
        \end{itemize}
        
    \item \textbf{coplanar front}: Coplanar in front.
        \begin{itemize}
            \item Determine if the front edges of the two objects are coplanar.
        \end{itemize}
        
    \item \textbf{coplanar back}: Coplanar in the back.
        \begin{itemize}
            \item Determine if the back edges of the two objects are coplanar.
        \end{itemize}
    \end{enumerate}

\section{Objects Generated With Workflow}
\begin{figure}[ht!]
    \centering
    \includegraphics[width=0.45\textwidth]{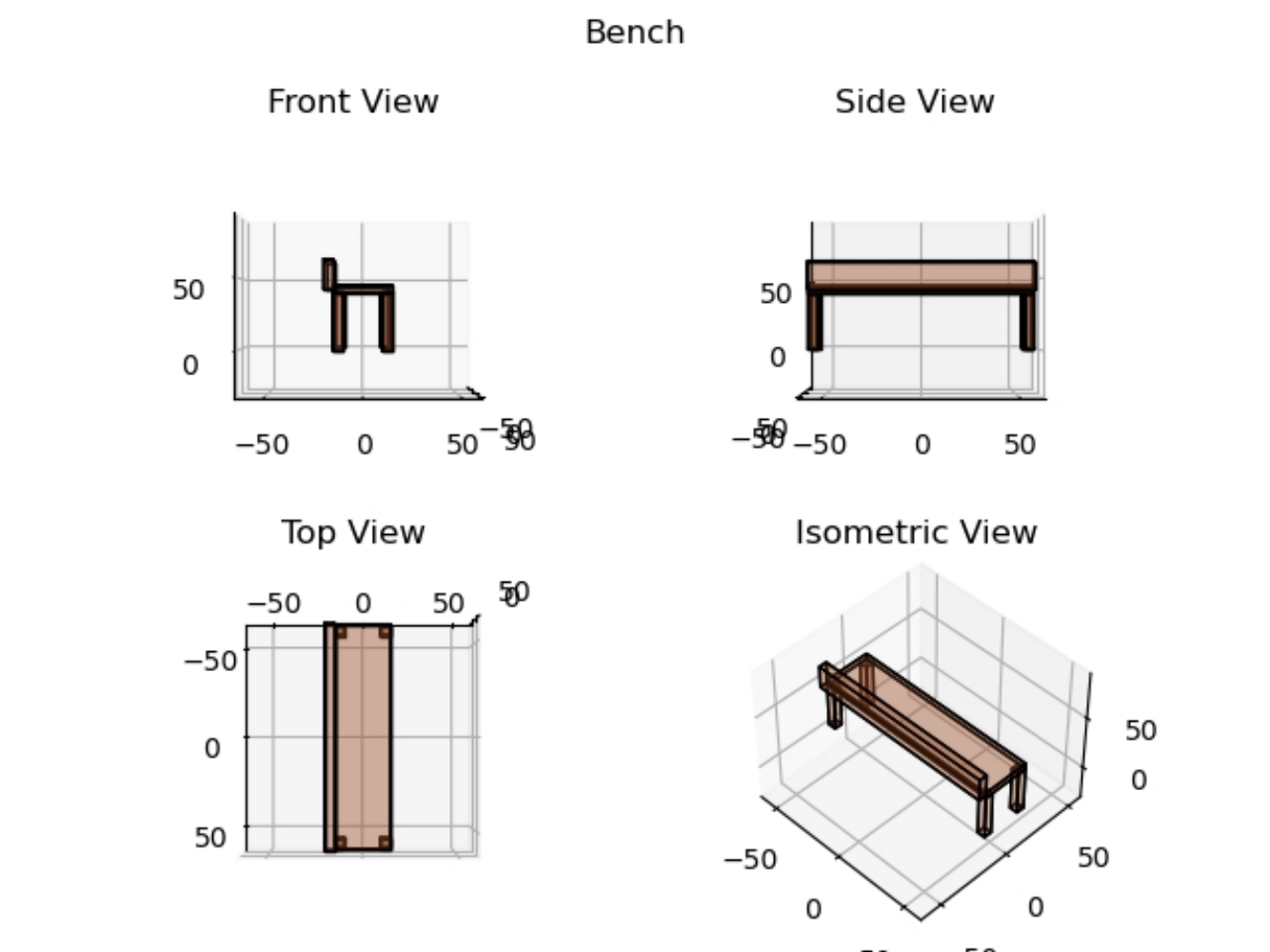}
    \caption{Bench}
    \label{fig:bench}
\end{figure}
\hfill
\begin{figure}[ht!]
    \centering
    \includegraphics[width=0.45\textwidth]{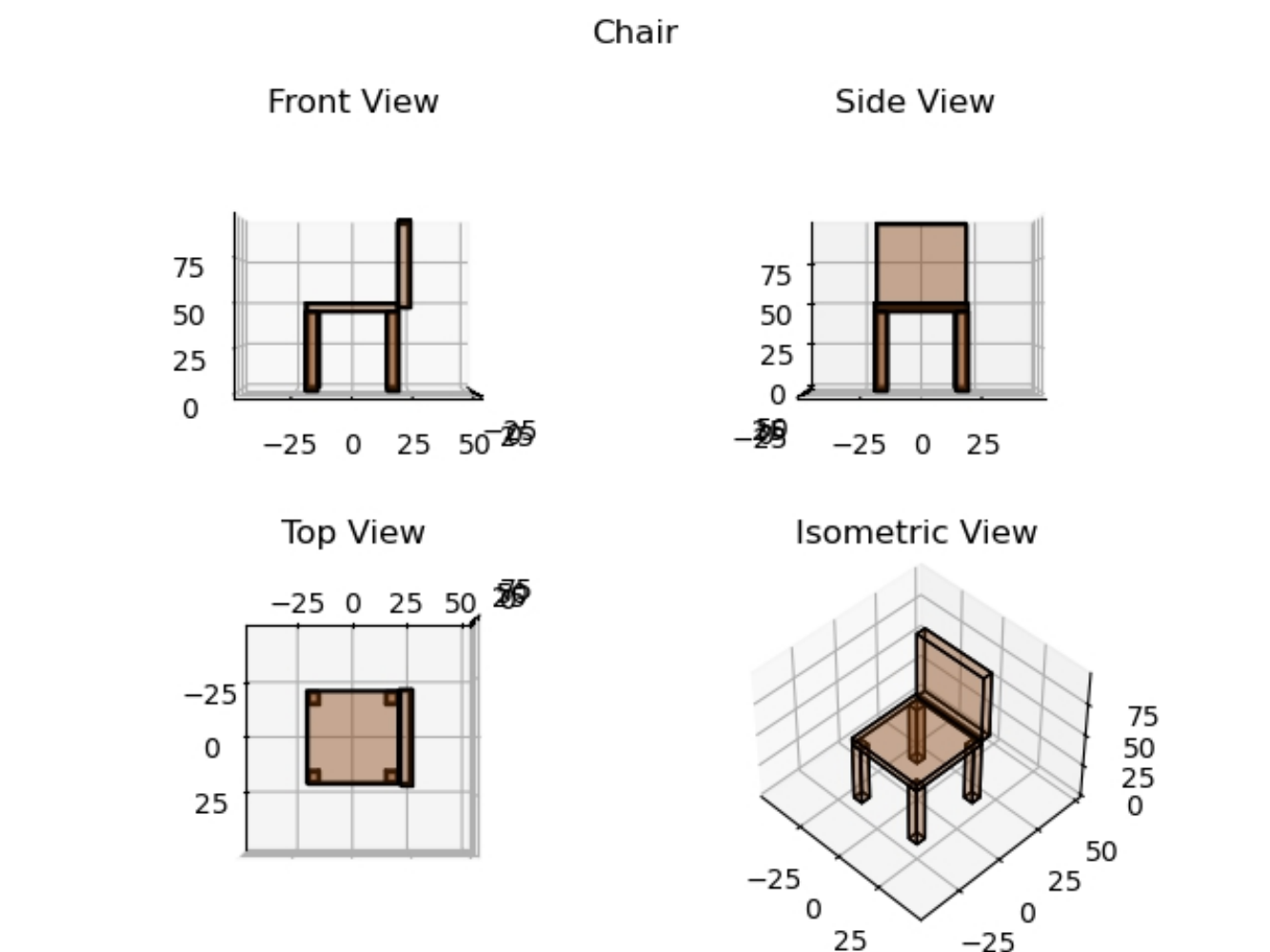}
    \caption{Chair}
    \label{fig:chair}
\end{figure}
\hfill
\begin{figure}[ht!]
    \centering
    \includegraphics[width=0.45\textwidth]{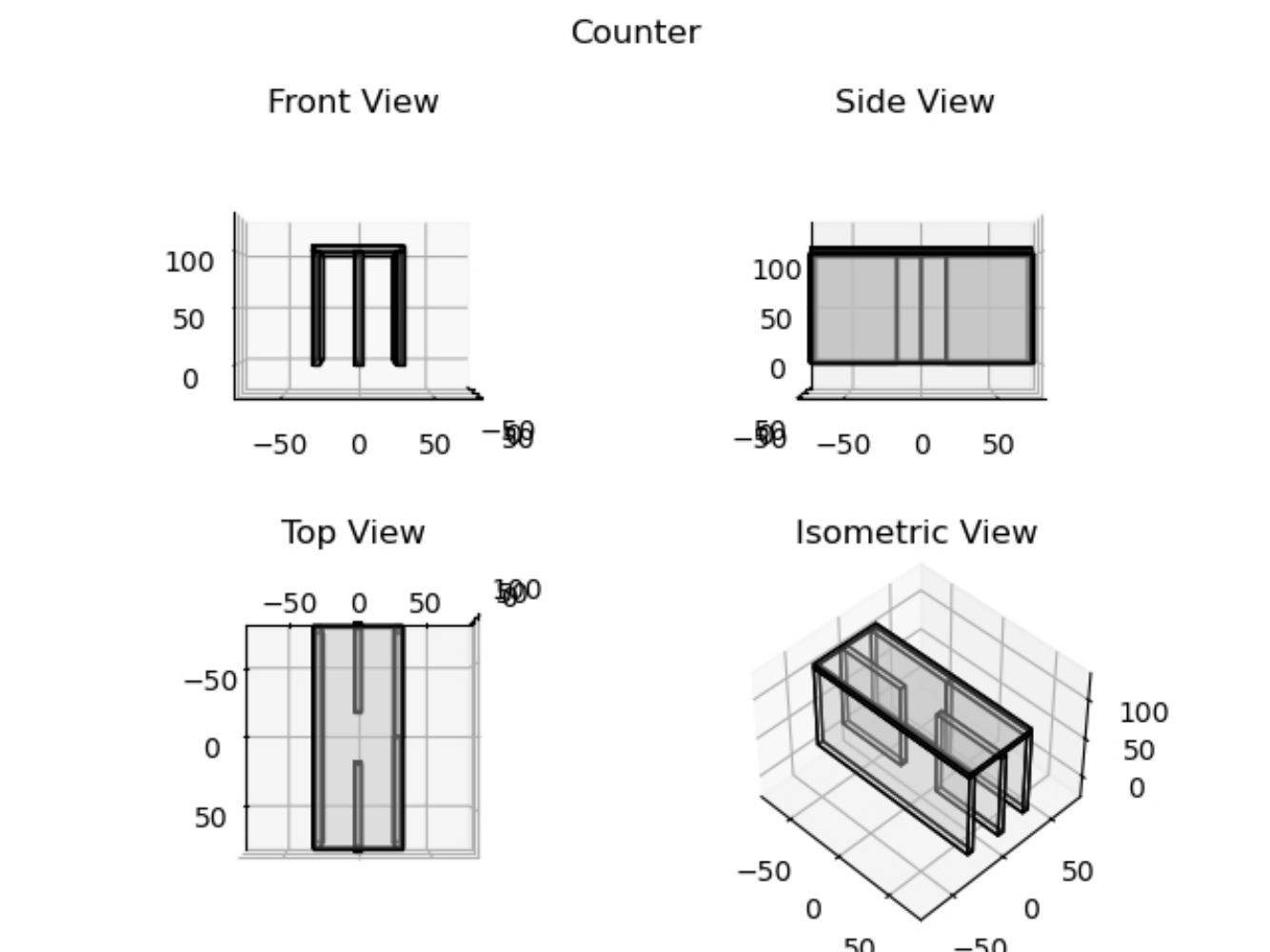}
    \caption{Counter}
    \label{fig:counter}
\end{figure}
\hfill
\begin{figure}[ht!]
    \centering
    \includegraphics[width=0.45\textwidth]{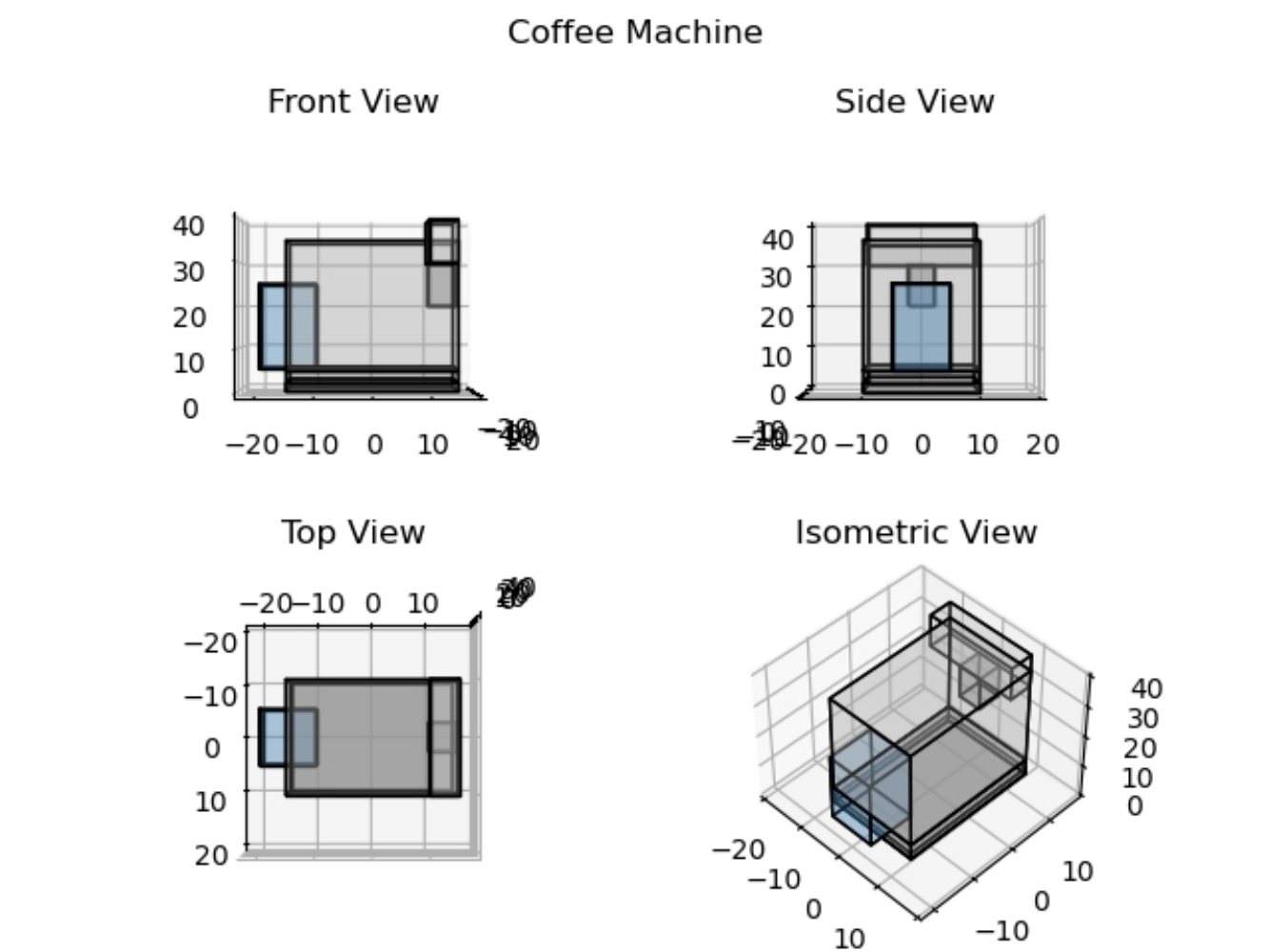}
    \caption{Coffee Machine}
    \label{fig:coffee_machine}
\end{figure}
\hfill
\begin{figure}[ht!]
    \centering
    \includegraphics[width=0.45\textwidth]{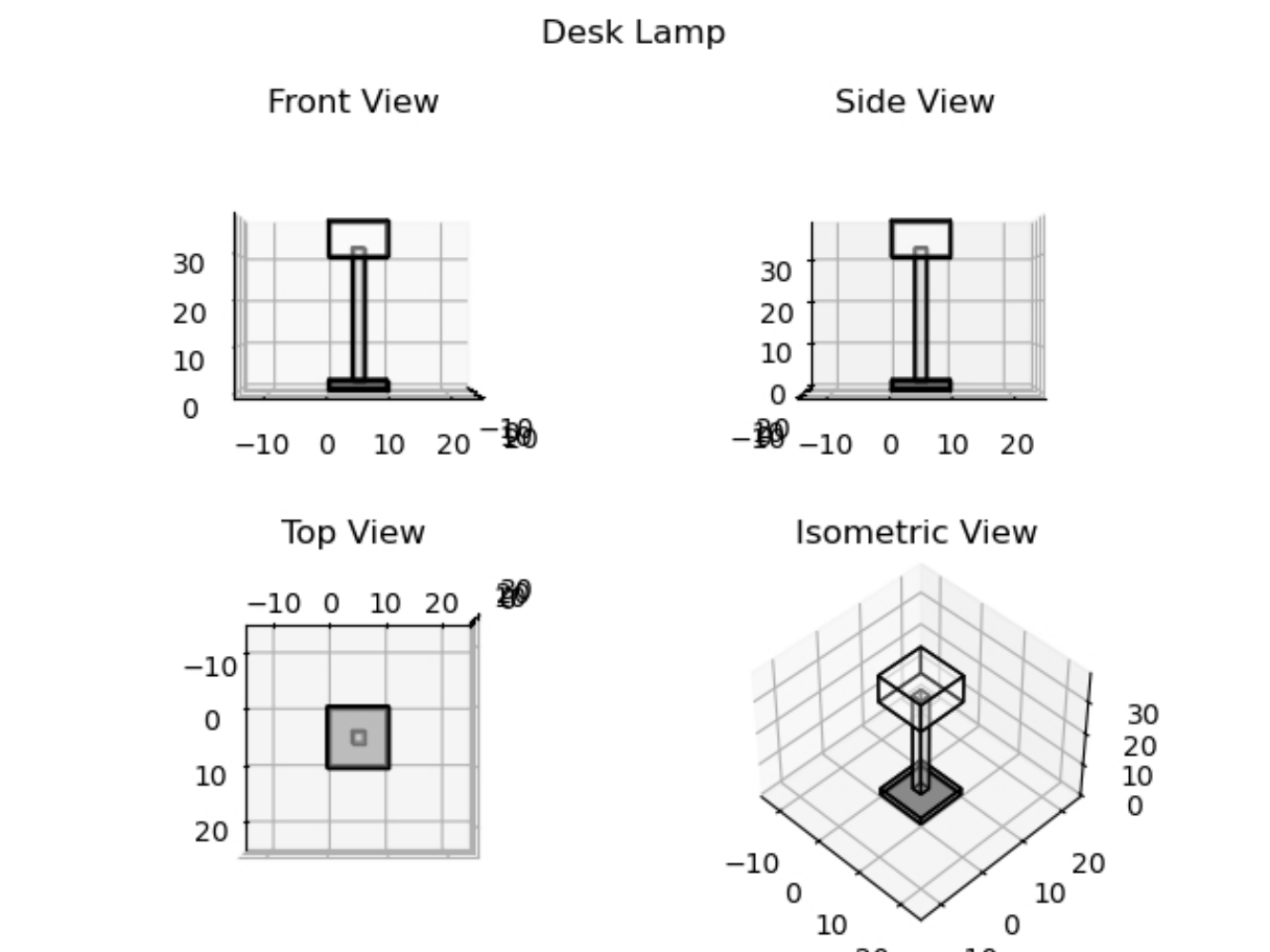}
    \caption{Lamp}
    \label{fig:lamp}
\end{figure}
\hfill
\begin{figure}[ht!]
    \centering
    \includegraphics[width=0.45\textwidth]{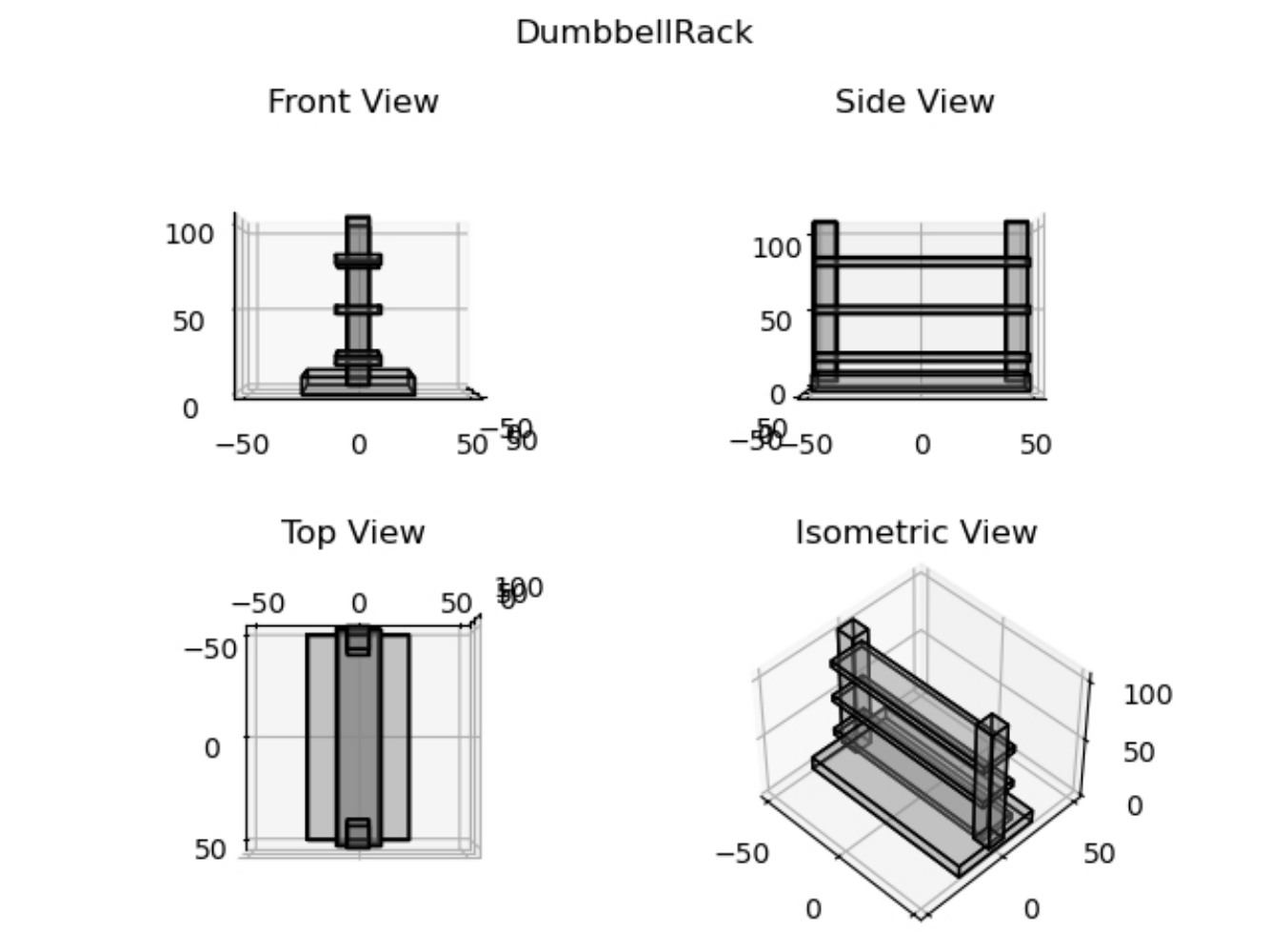}
    \caption{Dumbbell Rack}
    \label{fig:dumbbell_rack}
\end{figure}
\hfill
\begin{figure}[ht!]
    \centering
    \includegraphics[width=0.45\textwidth]{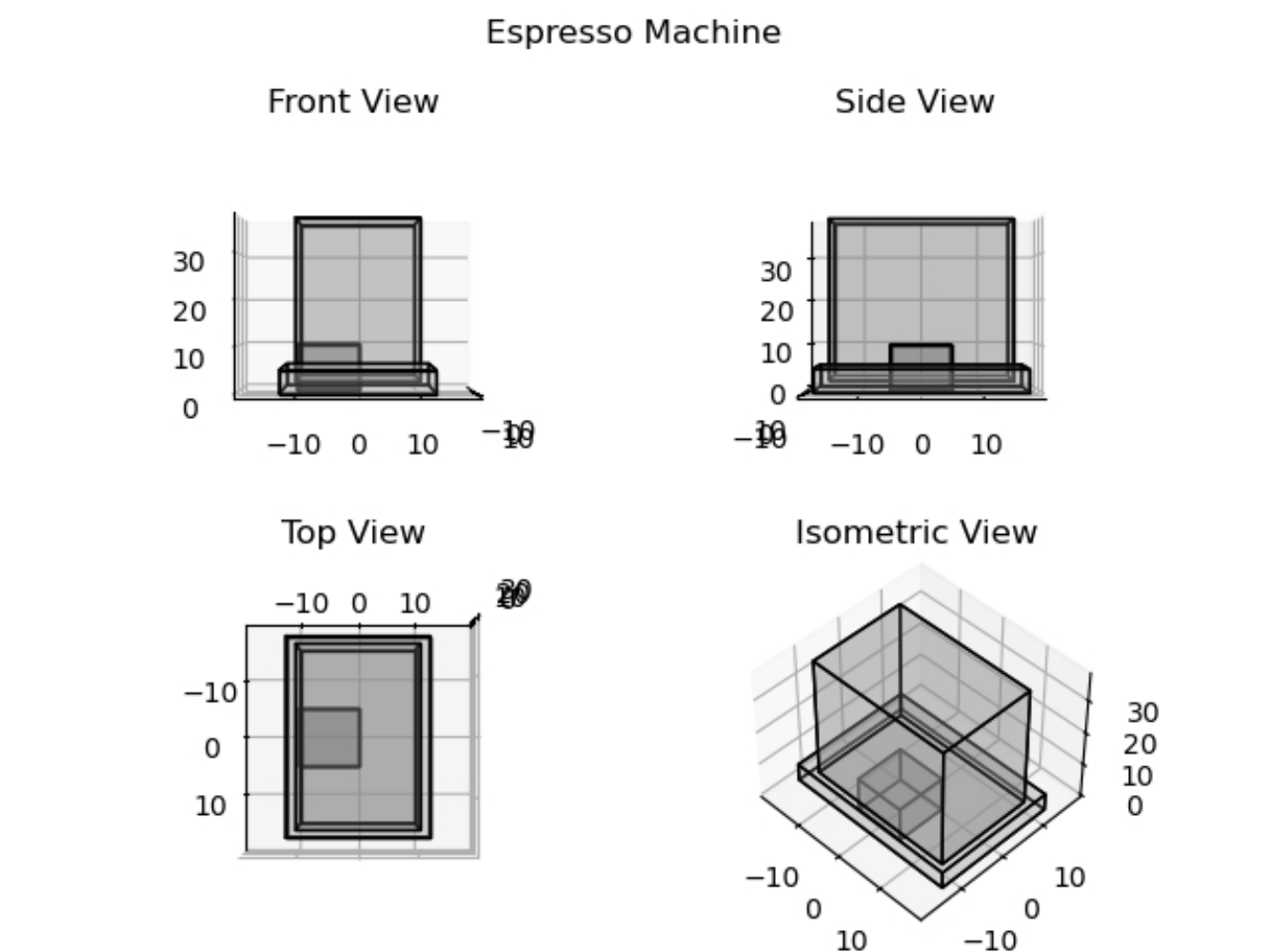}
    \caption{Espresso Machine}
    \label{fig:espresso_machine}
\end{figure}
\hfill
\begin{figure}[ht!]
    \centering
    \includegraphics[width=0.45\textwidth]{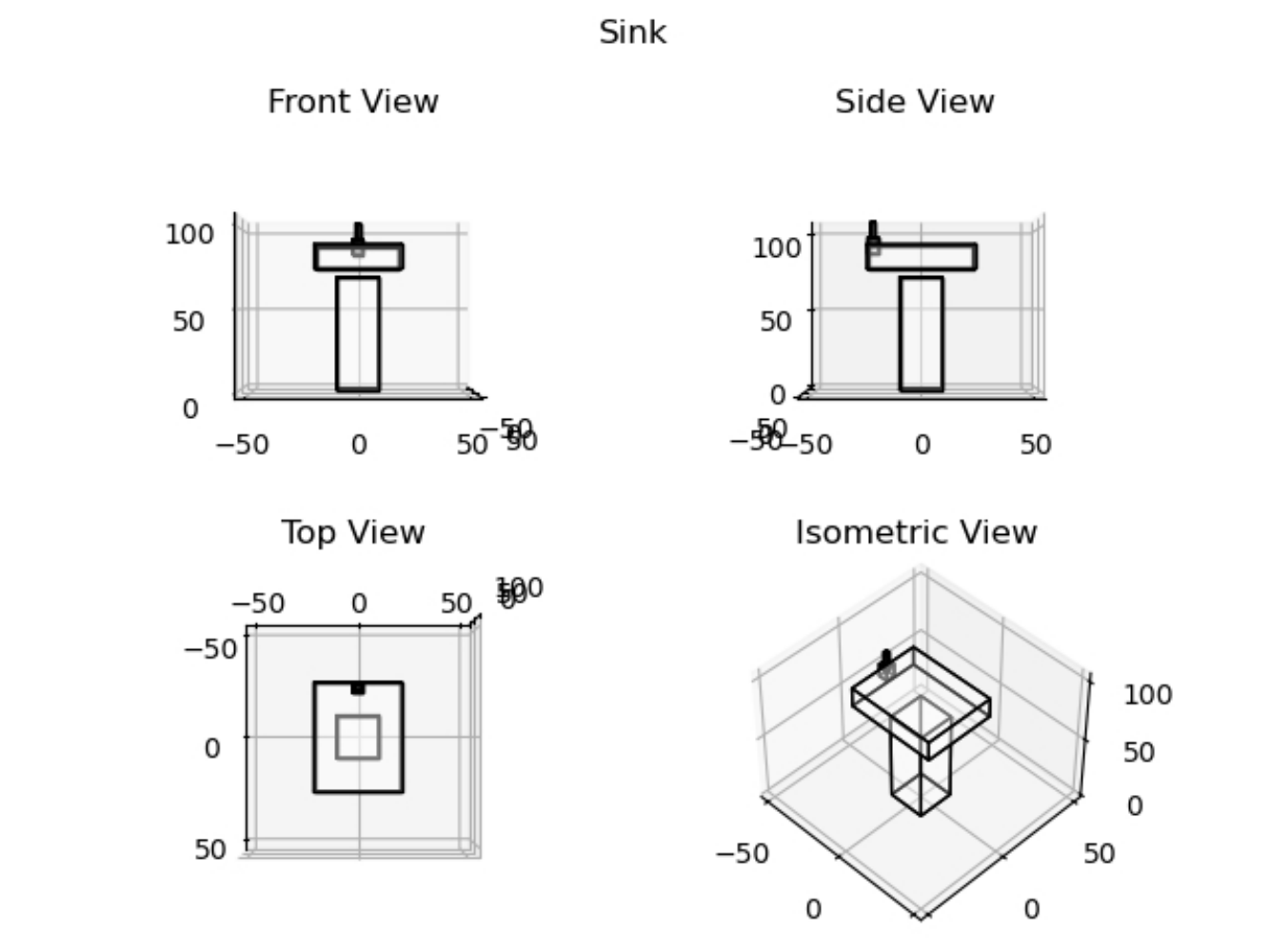}
    \caption{Sink}
    \label{fig:sink}
\end{figure}
\hfill
\begin{figure}[ht!]
    \centering
    \includegraphics[width=0.45\textwidth]{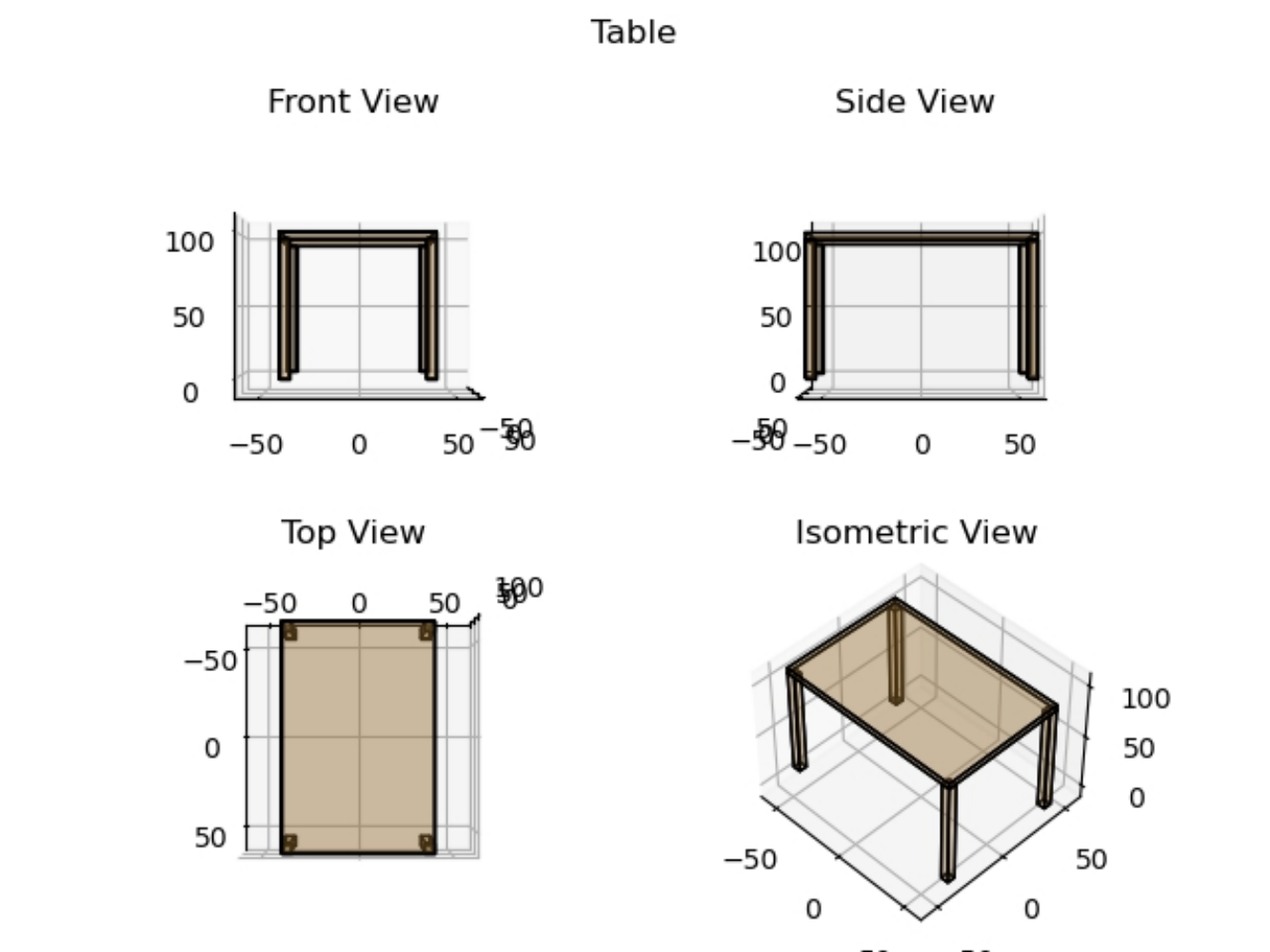}
    \caption{Table}
    \label{fig:table}
\end{figure}
\hfill
\begin{figure}[ht!]
    \centering
    \includegraphics[width=0.45\textwidth]{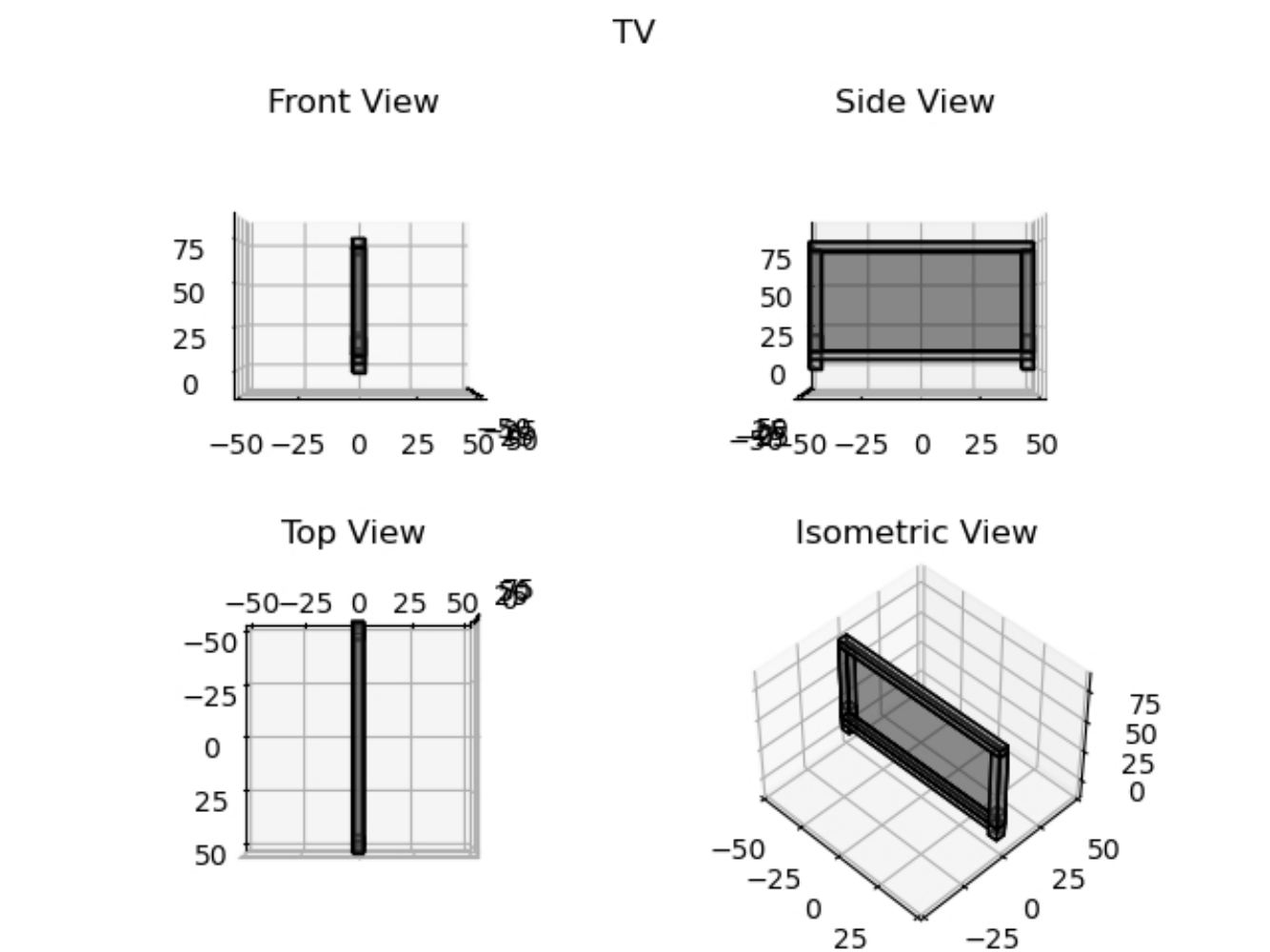}
    \caption{TV}
    \label{fig:tv}
\end{figure}
\hfill
\begin{figure}[ht]
    \centering
    \includegraphics[width=0.45\textwidth]{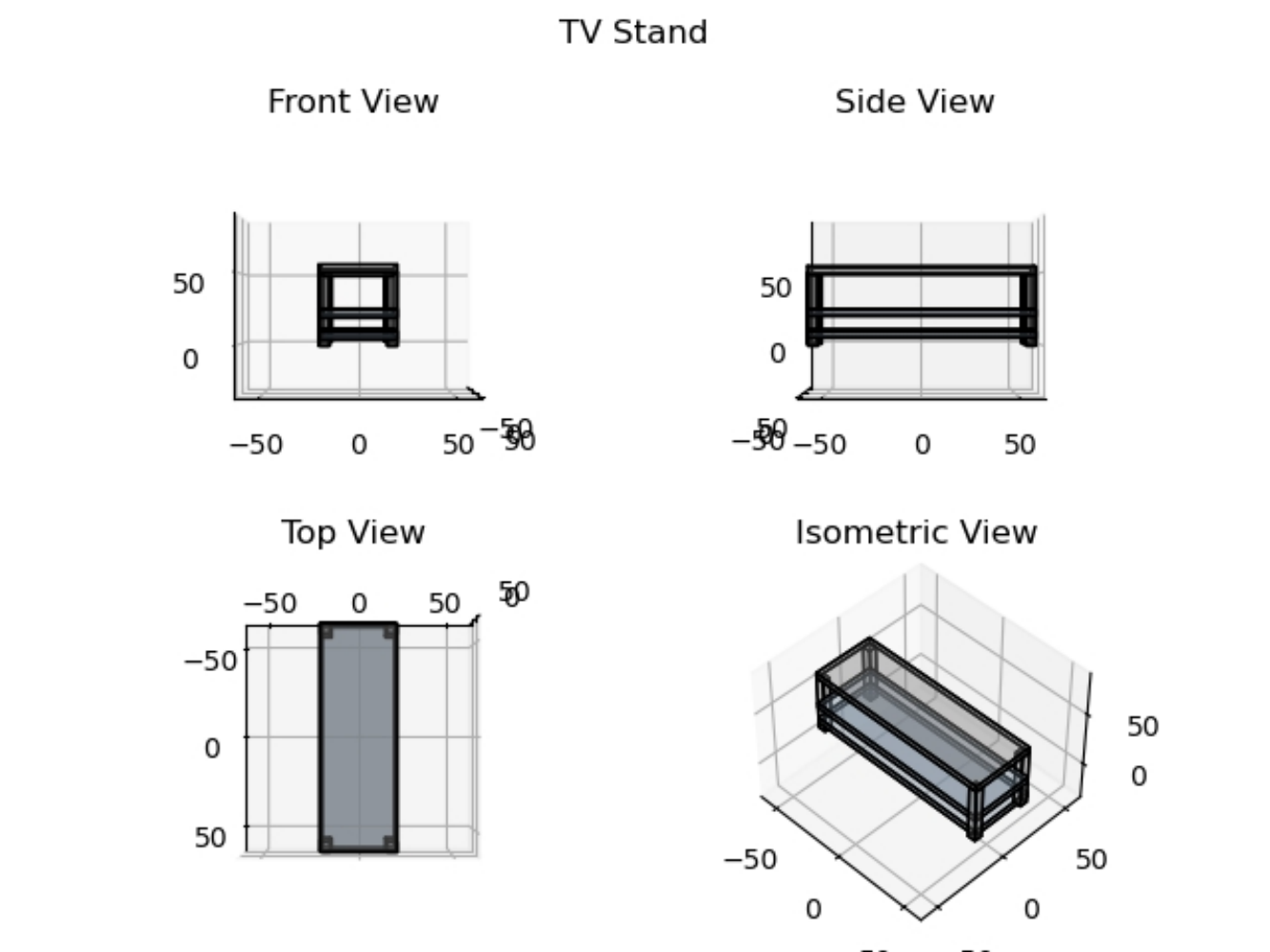}
    \caption{TV Stand}
    \label{fig:tv_stand}
\end{figure}
\label{fig:objects example}

\end{document}